\pdfoutput=1
\documentclass[11pt]{article}
\usepackage[preprint]{style}
\usepackage{times}
\usepackage{latexsym}
\usepackage[T1]{fontenc}
\usepackage{url}
\usepackage[utf8]{inputenc}
\usepackage{microtype}
\usepackage{inconsolata}
\usepackage{amsmath}
\usepackage{amsthm}
\usepackage{algorithm}
\usepackage{algorithmic}
\usepackage[switch]{lineno}
\usepackage{enumitem}
\usepackage{graphicx}
\usepackage{adjustbox}
\usepackage{amssymb}
\usepackage{pifont}
\usepackage{caption}
\usepackage{color}
\usepackage{multirow} 
\usepackage{threeparttable}
\usepackage{colortbl} 
\usepackage{booktabs}
\usepackage{indentfirst}
\usepackage[normalem]{ulem}
\DeclareUnicodeCharacter{0394}{\ensuremath{\Delta}}
\usepackage[normalem]{ulem}
\useunder{\uline}{\ul}{}
\usepackage{pifont}
\newcommand{\cmark}{\textcolor{blue}{\checkmark}}
\newcommand{\xmark}{\textcolor{black}{\ding{55}}}

\title{Shortened LLaMA: Depth Pruning for Large Language Models\\with Comparison of Retraining Methods}

\author{
\begin{tabular}{@{}c c}
Bo-Kyeong Kim$^{1}$\thanks{Equal contribution.}\quad
Geonmin Kim$^{1}$\footnotemark[1]\quad
Tae-Ho Kim$^{1}$\thanks{Corresponding author.}\quad\\
Thibault Castells$^{1}$\quad
Shinkook Choi$^{1}$\quad
Junho Shin$^{1}$\quad
Hyoung-Kyu Song$^{2}$\vspace{6pt} 
\end{tabular}\\
$^{1}$Nota Inc.\quad$^{2}$Captions\\
\texttt{\scalebox{0.715}{ \{bokyeong.kim, geonmin.kim, thkim, thibault, shinkook.choi, junho.shin\}@nota.ai, kyu@captions.ai} }
}

\begin{document}
\maketitle
\begin{abstract}
Structured pruning of modern large language models (LLMs) has emerged as a way of decreasing their high computational needs. Width pruning reduces the size of projection weight matrices (e.g., by removing attention heads) while maintaining the number of layers. Depth pruning, in contrast, removes entire layers or blocks, while keeping the size of the remaining weights unchanged. Most current research focuses on either width-only or a blend of width and depth pruning, with little comparative analysis between the two units (width~\textit{vs.}~depth) concerning their impact on LLM inference efficiency. In this work, we show that simple depth pruning can effectively compress LLMs while achieving comparable or superior performance to recent width pruning studies. Our pruning method boosts inference speeds, especially under memory-constrained conditions that require limited batch sizes for running LLMs, where width pruning is ineffective. In retraining pruned models for quality recovery, continued pretraining on a large corpus markedly outperforms LoRA-based tuning, particularly at severe pruning ratios. We hope this work can help build compact yet capable LLMs.
\end{abstract}

\section{Introduction}

The advancement of large language models (LLMs)~\cite{touvron2023llama,openai2023gpt4,chowdhery2022palm,zhang2022opt,scao2022bloom} has brought significant improvements in language-based tasks, enabling versatile applications such as powerful chatbots~\cite{bard,chatgpt}. However, the deployment of LLMs is constrained by their intensive computational demands. To make LLMs more accessible and efficient for practical use, various optimization strategies have been actively studied over recent years (see~\citet{zhu2023survey,wan2023efficient} for survey). This work focuses on \textit{structured} pruning~\cite{fang2023depgraph,li2017pruning}, which removes groups of unnecessary weights and can facilitate hardware-agnostic acceleration. 

\begin{figure}[t]
  \centering
    \includegraphics[width=\linewidth]{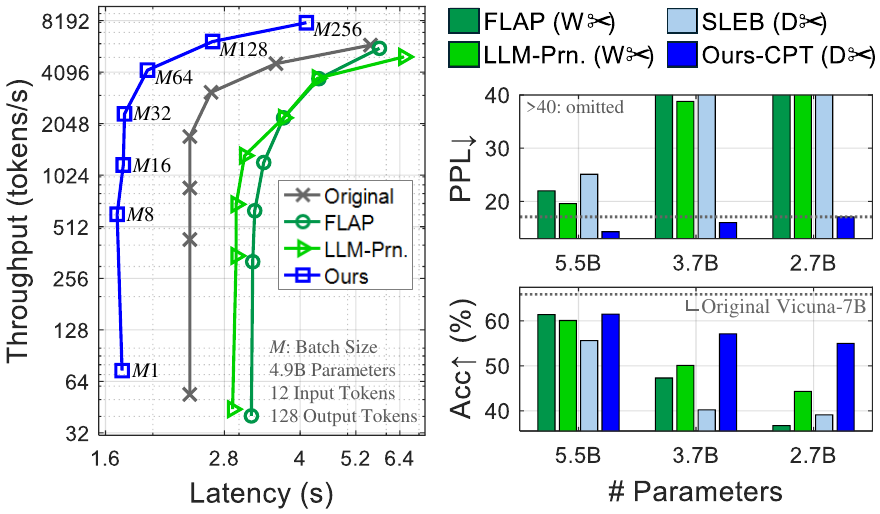}
        \vspace{-0.25in}
  \caption{Inference of pruned Vicuna-7B models on an NVIDIA H100 GPU. \uline{Left}: Compared to width pruning (W\ding{34}) of FLAP~\cite{flap} and LLM-Pruner~\cite{llmpruner}, our depth pruning (D\ding{34}) achieves faster inference. \uline{Right}: Continued pretraining is crucial for restoring the quality of heavily pruned models with fewer than 3.7B parameters, enabling our method to surpass the baselines, including SLEB~\cite{song2024sleb}. See Table~\ref{table:cpt_results} for details.} \label{fig_teaser}
  \vspace{-0.1in}
\end{figure}

\begin{figure*}[t]
  \centering
    \includegraphics[width=\linewidth]{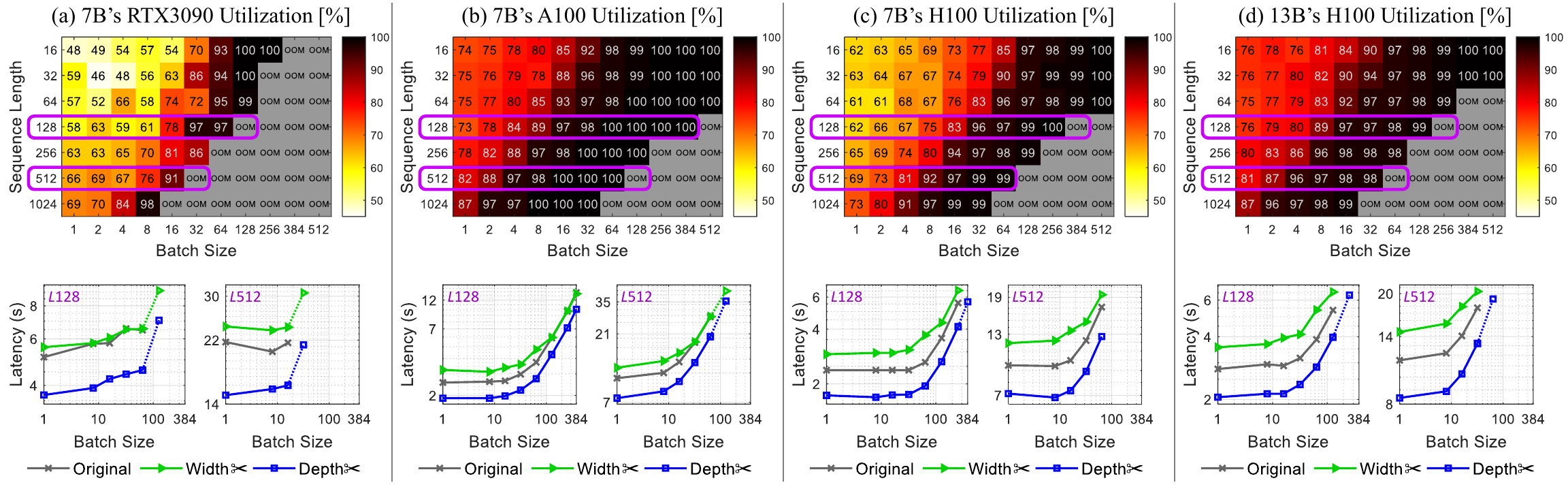}
        \vspace{-0.23in}
  \caption{\uline{Top}: GPU compute utilization of (a)–(c) running LLaMA-7B on different NVIDIA GPUs and that of (d) Vicuna-13B. Increasing batch sizes can enhance GPU utilization and throughput, but pushing this too far triggers OOM issues. \uline{Bottom}: Latency results ($L$: target output length). Our depth pruning (blue lines) improves generation speeds over the original models (gray), while width pruning~\cite{llmpruner} is ineffective (green). The dotted lines show that pruned models can operate with larger batch sizes that cause OOM errors for the original model. The results are obtained with pruning ratios of 27\% for the 7B model and 29\% for the 13B model.
  }
  \vspace{-0.1in}
  \label{fig_gpuutil}
\end{figure*}

In the context of compressing recent LLMs, LLM-Pruner~\cite{llmpruner} and FLAP~\cite{flap} narrow the network width by pruning coupled structures (e.g., attention heads and their associated weight connections) while maintaining the number of layers. Sheared-LLaMA~\cite{xia2023sheared} reduces not only the network width but also its depth by entirely removing some layers. Despite the existence of pruning methods~\cite{xia2022structured,kurtic2023ziplm,xia2023sheared} that incorporate both width and depth aspects, there remains a gap in detailed analysis comparing these two factors (width~\textit{vs.}~depth), specifically in relation to their impact on LLM inference efficiency.

In addition to substantial model sizes, LLM inference is distinguished by an autoregressive decoding mechanism, which predicts tokens one by one based on the input and the previously generated tokens. This sequential generation process often exhibits a memory-bound nature, leading to considerable underutilization of GPU compute abilities~\cite{kwon2023efficient,jin2023s}. While expanding batch sizes is a standard way to enhance GPU utilization and throughput, this approach is unfeasible for low-specification GPUs with memory constraints. We aim to improve inference speeds of LLMs, especially under hardware limitations that demand small batch sizes, where we observe that width-only pruning is inadequate.

Depth pruning is often regarded as being less effective in generation performance compared to width pruning, due to the elimination of bigger and coarse units. Contrary to the prevailing view, this study reveals that depth pruning is a compelling option for compressing LLMs, and it can achieve comparable or superior performance to prior studies depending on the retraining setups. Our contributions are summarized as follows:

\begin{enumerate}[itemsep=0em]
\setlength{\leftskip}{-0.22cm}
\vspace{-0.025in}
\item[$\circ$] In scenarios with limited batch sizes, our work demonstrates that width pruning is difficult to attain actual speedups in LLM's autoregressive generation. This aspect has been underexplored in previous works.

\vspace{-0.02in}
\item[$\circ$] We introduce a simple yet effective method for depth pruning of LLMs by exploring various design factors. Our compact LLMs, obtained by excluding several Transformer blocks, achieve actual speedups. 

\vspace{-0.02in}
\item[$\circ$] We show that under moderate pruning ratios, our depth pruning method with LoRA retraining can rival recent width pruning studies for LLMs in zero-shot capabilities. For more aggressive pruning (over 40\% removal), intensive retraining with a full-parameter update is crucial for recovering performance.

\setlength{\leftskip}{0pt}
\end{enumerate}

\begin{figure}[t]
  \centering
    \includegraphics[width=\linewidth]{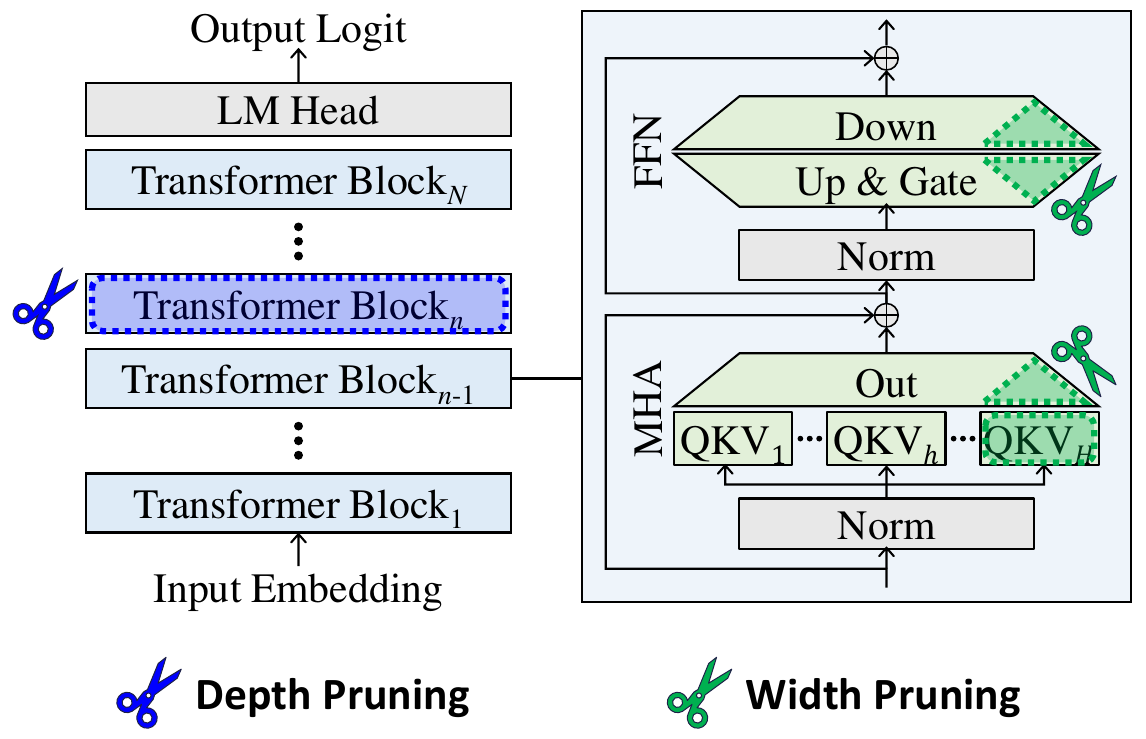}
        \vspace{-0.25in}
    \caption{Comparison of pruning units. Width pruning reduces the size of projection weight matrices. Depth pruning removes Transformer blocks, or individual MHA and FFN modules.}\label{fig_compare_depth_width_prune}  
  \vspace{-0.1in}
\end{figure}

\vspace{-0.2in}
\section{Problem: Small-batch LLM Inference}

Most LLMs are autoregressive models that sequentially produce tokens, based on the initial prompt and the sequence of tokens previously generated. The token-by-token generation process often involves multiplying large matrices (weights) with smaller matrices or vectors (activations). The primary bottleneck for inference efficiency is memory access operations rather than the speed of mathematical computations (referred to as `memory-bound'), leading to suboptimal use of GPU computing power~\cite{kwon2023efficient}. Though increasing batch sizes is a standard way to enhance GPU computation and throughput, it poses a risk of out-of-memory (OOM) errors (see Figure~\ref{fig_gpuutil})\footnote{Using the HF-Transformers library~\cite{wolf-etal-2020-transformers}, we ran the LLMs with 12 input tokens for 20 batched runs after 10 warm-ups. Top: Peak GPU compute utilization~\cite{nvidia_smi_query}. Bottom: Mean latency over 20 runs.} unless advanced system-level optimizations~\cite{kwon2023efficient,sheng2023flexgen} are applied.

In this study, our focus is on accelerating the inference of LLMs under small-batch conditions caused by hardware restrictions. Such situations are relevant for deploying LLMs on memory-constrained local devices, which can enhance user experience and data privacy protection. We show that (i) reducing weight shapes via width pruning does not improve generation speeds and can even degrade it when the resulting weight dimensions are unsuitable for GPU capabilities, and (ii) notable speed gains are only achievable through depth pruning that excludes a number of modules entirely.

\section{Method: Block Pruning} \label{method}

An LLM is a stack of multiple Transformer blocks \cite{transformer}, each of which contains a pair of multi-head attention (MHA) and feed-forward network (FFN) modules (see Figure~\ref{fig_compare_depth_width_prune}). We choose this Transformer block as the prunable unit to prioritize reducing inference latency. Our approach is simple: after identifying unimportant blocks with straightforward metrics, we perform simple one-shot pruning.

\subsection{Evaluation of Block-level Importance} \label{subsect_crit}
We consider the following criteria to evaluate the significance of each block, ultimately selecting the Taylor+ and PPL metrics (see Table~\ref{table:criterion}). Specifically, the linear weight matrix is denoted as $\mathbf{W}^{k,n} = \left[W_{i,j}^{k,n}\right]$ with a size of $(d_{\mathrm{out}}, d_{\mathrm{in}})$, where $k$ represents the type of operation (e.g., a query projection in MHA or an up projection in FFN) within the $n$-th Transformer block. The weight importance scores are calculated at the output neuron level~\cite{wanda}, followed by summing\footnote{In our exploration of various aggregation strategies (i.e., sum, mean, product, and max operations), summing the scores was effective at different pruning ratios.} these scores to assess the block-level importance.
 
\paragraph{Magnitude (Mag).} This metric~\cite{li2016pruning} is a fundamental baseline in the pruning literature, assuming that weights with smaller norms are less informative. For the block-level analysis, we compute $I_{\mathrm{Magnitude}}^n = \sum_k \sum_i \sum_j \left| W_{i,j}^{k,n} \right|$.

\begin{figure}[t]
\centering
\includegraphics[width=0.9\linewidth]{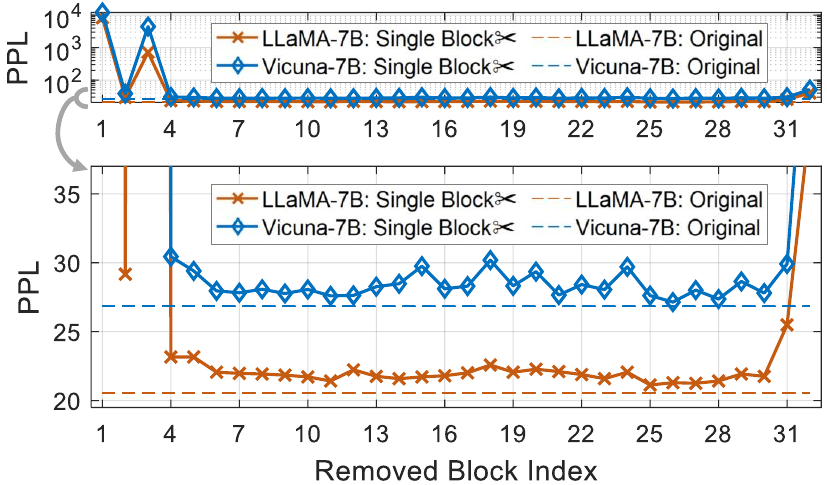}
\vspace{-0.05in}
\caption{Estimated importance of each Transformer block on the calibration set. We prune blocks that have lower (better) PPL scores, as their removal causes less disruption to the output.}\label{fig_ppl_crit}
\vspace{-0.1in}
\end{figure}

\paragraph{Taylor.} Assessing the error caused by the removal of a weight parameter helps in identifying its significance. For a given calibration dataset $D$, this can be expressed as the alteration in the training loss $\mathcal{L}$~\cite{lecun1989optimal,molchanov2019importance}: $\left| \mathcal{L}(W_{i,j}^{k,n}; D) - \mathcal{L}(W_{i,j}^{k,n} = 0; D) \right| \approx \left| \frac{\partial \mathcal{L}(D)}{\partial W_{i,j}^{k,n}} W_{i,j}^{k,n} \right|$, where we omit the second-order derivatives by following~\citet{llmpruner}. We define the block score as $I_{\mathrm{Taylor}}^n = \sum_k \sum_i  \sum_j \left| \frac{\partial \mathcal{L}(D)}{\partial W_{i,j}^{k,n}} W_{i,j}^{k,n} \right|$.

\paragraph{Mag+ and Taylor+.} Upon using the aforementioned metrics, the early blocks are labeled as unimportant, but their removal leads to severe performance drops. Similar to a popular heuristic~\cite{gale2019state,lee2021layeradaptive}, we preserve the first four and the last two blocks~\cite{llmpruner} by excluding them from the pruning candidates.

\paragraph{Perplexity (PPL).} Redundant blocks contribute less to the model's outputs, and their removal leads to smaller degradation in PPL, a commonly used metric for language modeling tasks. In this context, we eliminate each block from the source model and monitor its influence on PPL using the calibration set $D$: $I_{\mathrm{PPL}}^n = \exp \left\{ -\frac{1}{SL} \sum_{s} \sum_{l} \log p_{\theta^{n}}(x_{l}^{(s)} | x_{<l}^{(s)}) \right\}$, where $\theta^{n}$ denotes the model without its $n$-th block, and $s = 1, \ldots, S$ and $l = 1, \ldots, L$ are the indices for sequences and tokens in $D$. The PPL can be derived from the next-token prediction loss and requires only forward-pass computation. As shown in Figure~\ref{fig_ppl_crit}, several blocks are removable with only a slight effect on the PPL metric. Pruning initial and final blocks significantly degrades the performance, which necessitates keeping them unpruned. 

\subsection{One-shot Pruning}
After sorting the block-level importance scores, we prune the less crucial blocks in a single step. Since every block has an identical configuration and it is easy to calculate the number of parameters for one block, we readily decide how many blocks should be removed to meet the target model size.

Iterative pruning with intermediate updates of block importance can be applied as in SLEB~\cite{song2024sleb}. However, it requires much longer computing time than one-shot pruning as the number of blocks increases. Furthermore, we empirically observed that retraining strategies matter more than whether the pruning scheme is iterative or one-shot, especially under severe pruning ratios.

\subsection{Retraining for Performance Restoration}
Some recent studies suggest that structured pruning of LLMs can be retraining-free~\cite{song2024sleb,flap} or feasible with low retraining budgets~\cite{llmpruner}. However, the types of retraining over different pruning rates have been underexplored. Here, we compare several retraining strategies and their implications for regaining the quality of pruned models.

\paragraph{Low-Rank Adaptation (LoRA).} LoRA~\cite{lora} enables the efficient refinement of LLMs with less computation. \citet{llmpruner} has applied LoRA to enhance moderately width-pruned models (e.g., with 20\% of units removed) on an instruction tuning dataset. In this work, we show that LoRA can also recover the ability of depth-pruned models; however, it does not perform well for extensive compression rates (e.g., with over 50\% removal) in either width or depth pruning.

\paragraph{Continued Pretraining (CPT).} We leverage CPT, which involves updating all parameters, on a large-scale pretraining corpus. This powerful retraining is critical for severely depth-pruned models, extending its proven effectiveness for width- or hybrid-pruned models~\cite{xia2023sheared}. Though requiring greater resources than LoRA, CPT on pruned networks significantly accelerates learning and yields superior results compared to training the same architectures from random initialization.

\paragraph{CPT$\Rightarrow$LoRA} Once CPT on the pretraining data is completed, LoRA with the instruction set is applied to observe whether further performance improvement can be achieved.

\begin{table}[t]
\centering
\begin{adjustbox}{max width=\columnwidth}
\begin{threeparttable}
\begin{tabular}{cc|c|c|cc}
\specialrule{.2em}{.1em}{.1em} 

\multicolumn{2}{c|}{Model}         & \#Param & \#Block\textsuperscript{$\ddagger$} & \#Head\textsuperscript{$\ddagger$}    & FFN-D\textsuperscript{$\ddagger$}          \\ \hline
\multicolumn{2}{c|}{Original 7B}   & 6.7B    & 32      & 32        & 11008          \\ \hline
\multirow{4}{*}{35\%\textsuperscript{$\dagger$}} & Wanda-sp   & 4.5B    & 32      & 21        & 7156           \\
                      & FLAP       & 4.5B    & 32      & 23.0{\scriptsize±8.8}  & 6781.1{\scriptsize±2440.6}  \\
                      & LLM-Pruner & 4.4B    & 32      & 18        & 6054           \\ \cline{2-6} 
                      & Ours       & 4.5B    & 21      & 32        & 11008          \\

\specialrule{.2em}{.1em}{.1em}
\specialrule{.2em}{.1em}{.1em}

\multicolumn{2}{c|}{Original 13B}  & 13.0B   & 40      & 40        & 13824          \\ \hline
\multirow{4}{*}{37\%\textsuperscript{$\dagger$}} & Wanda-sp   & 8.4B    & 40      & 26        & 8710           \\
                      & FLAP       & 8.3B    & 40      & 27.5{\scriptsize±11.3} & 8326.6{\scriptsize±2874.9}  \\
                      & LLM-Pruner & 8.2B    & 40      & 22        & 7603           \\ \cline{2-6} 
                      & Ours       & 8.3B    & 25      & 40        & 13824          \\ 
                      
\specialrule{.2em}{.1em}{.1em} 

\end{tabular}
\begin{tablenotes}[para,flushleft]
\footnotesize 
\textsuperscript{$\dagger$}Reduction ratio for the number of parameters.
\newline
\textsuperscript{$\ddagger$}\#Block: \#Transformer blocks; \#Head: \#attention heads of MHA; FFN-D: intermediate size of FFN. 
\end{tablenotes}
\end{threeparttable}
\end{adjustbox}
\vspace{-0.05in}
\caption{Examples of pruned architectures on 7B-parameter (top) and 13B-parameter (bottom) models. While Wanda-sp~\cite{wanda,flap}, FLAP~\cite{flap}, and LLM-Pruner~\cite{llmpruner} reduce the network width, our method reduces the network depth. See Table~\ref{supple_table:arch} for the details.}
\label{table:arch_short_ver}

\vspace{-0.1in}
\end{table}

\begin{table*}[t]
\centering
\begin{adjustbox}{max width=0.92\linewidth}
\begin{threeparttable}
\begin{tabular}{ccc|ccc|cc|cc}
\specialrule{.2em}{.1em}{.1em} 

\multicolumn{3}{c|}{}                                                                                                                              & \multicolumn{3}{c|}{Zero-shot Performance}                                                                                                                                     & \multicolumn{2}{c|}{H100 80GB\textsuperscript{$\ddagger$}}                                                                                                                                & \multicolumn{2}{c}{RTX3090 24GB\textsuperscript{$\ddagger$}}                                                                                                                              \\ \cline{4-10} 
\multicolumn{3}{c|}{}                                                                                                                              & \multicolumn{2}{c|}{PPL↓}                                                                          &                                                                           &                                                                          &                                                                                    &                                                                          &                                                                                    \\
\multicolumn{3}{c|}{\multirow{-3}{*}{\#Param \& Method}}                                                                                                       & WikiText2                             & \multicolumn{1}{c|}{PTB}                                   & \multirow{-2}{*}{\begin{tabular}[c]{@{}c@{}}Ave Acc↑\\ (\%)\textsuperscript{$\dagger$}
\end{tabular}} & \multirow{-2}{*}{\begin{tabular}[c]{@{}c@{}}Latency↓\\ (s)\end{tabular}} & \multirow{-2}{*}{\begin{tabular}[c]{@{}c@{}}Throughput↑\\ (tokens/s)\end{tabular}} & \multirow{-2}{*}{\begin{tabular}[c]{@{}c@{}}Latency↓\\ (s)\end{tabular}} & \multirow{-2}{*}{\begin{tabular}[c]{@{}c@{}}Throughput↑\\ (tokens/s)\end{tabular}} \\ \hline
\multicolumn{3}{c|}{LLaMA-7B: 6.7B (Original)}                                                                                                               & 12.6                                  & \multicolumn{1}{c|}{22.1}                                  & 66.3                                                                      & 2.4                                                                      & 53.7                                                                               & 5.1                                                                      & 25.0                                                                               \\ \hline
                                                                                 &                         & Wanda-sp                              & 21.4                                  & \multicolumn{1}{c|}{47.2}                                  & 51.8                                                                      & 3.1                                                                      & 41.7                                                                               & 7.6                                                                      & 16.7                                                                               \\
                                                                                 &                         & FLAP                                  & \textbf{17.0}                         & \multicolumn{1}{c|}{\textbf{30.1}}                         & 59.5                                                                      & 3.2                                                                      & 40.5                                                                               & 7.7                                                                      & 16.5                                                                               \\
                                                                                 & \multirow{-3}{*}{W\ding{34}} & LLM-Pruner                            & 17.6                                  & \multicolumn{1}{c|}{30.4}                                  & 61.8                                                                      & 3.0                                                                      & 43.2                                                                               & 6.0                                                                      & 21.4                                                                               \\ \cline{2-10} 
                                                                                 &                         & SLEB                                  & 18.5                                  & \multicolumn{1}{c|}{31.6}                                  & 57.6                                                                      & \textbf{1.9}                                                             & \textbf{66.0}                                                                      & \textbf{4.5}                                                             & \textbf{28.4}                                                                      \\
                                                                                 &                         & \cellcolor[HTML]{ECF4FF}Ours: Taylor+ & \cellcolor[HTML]{ECF4FF}20.2          & \multicolumn{1}{c|}{\cellcolor[HTML]{ECF4FF}32.3}          & \cellcolor[HTML]{ECF4FF}\textbf{63.5}                                     & \cellcolor[HTML]{ECF4FF}\textbf{1.9}                                     & \cellcolor[HTML]{ECF4FF}\textbf{66.0}                                              & \cellcolor[HTML]{ECF4FF}\textbf{4.5}                                     & \cellcolor[HTML]{ECF4FF}\textbf{28.4}                                              \\
\multirow{-6}{*}{\begin{tabular}[c]{@{}c@{}}5.5B\\ (20\%\\  Pruned)\end{tabular}} & \multirow{-3}{*}{D\ding{34}} & \cellcolor[HTML]{ECF4FF}Ours: PPL     & \cellcolor[HTML]{ECF4FF}17.7          & \multicolumn{1}{c|}{\cellcolor[HTML]{ECF4FF}30.7}          & \cellcolor[HTML]{ECF4FF}61.9                                              & \cellcolor[HTML]{ECF4FF}\textbf{1.9}                                     & \cellcolor[HTML]{ECF4FF}\textbf{66.0}                                              & \cellcolor[HTML]{ECF4FF}\textbf{4.5}                                     & \cellcolor[HTML]{ECF4FF}\textbf{28.4}                                              \\ \hline
\hline
                                                                                 &                         & Wanda-sp                              & 133.6                                 & \multicolumn{1}{c|}{210.1}                                 & 36.9                                                                      & 3.1                                                                      & 41.6                                                                               & 8.0                                                                      & 16.1                                                                               \\
                                                                                 &                         & FLAP                                  & 25.6                                  & \multicolumn{1}{c|}{44.4}                                  & 52.7                                                                      & 3.2                                                                      & 40.5                                                                               & 8.1                                                                      & 15.8                                                                               \\
                                                                                 & \multirow{-3}{*}{W\ding{34}} & LLM-Pruner                            & 24.2                                  & \multicolumn{1}{c|}{40.7}                                  & \textbf{55.5}                                                             & 2.9                                                                      & 44.4                                                                               & 6.1                                                                      & 21.1                                                                               \\ \cline{2-10} 
                                                                                 &                         & SLEB                                  & 34.2                                  & \multicolumn{1}{c|}{49.8}                                  & 50.1                                                                      & \textbf{1.6}                                                             & \textbf{80.1}                                                                      & \textbf{3.4}                                                             & \textbf{37.8}                                                                      \\
                                                                                 &                         & \cellcolor[HTML]{ECF4FF}Ours: Taylor+ & \cellcolor[HTML]{ECF4FF}33.2          & \multicolumn{1}{c|}{\cellcolor[HTML]{ECF4FF}58.5}          & \cellcolor[HTML]{ECF4FF}55.4                                              & \cellcolor[HTML]{ECF4FF}\textbf{1.6}                                     & \cellcolor[HTML]{ECF4FF}\textbf{80.1}                                              & \cellcolor[HTML]{ECF4FF}\textbf{3.4}                                     & \cellcolor[HTML]{ECF4FF}\textbf{37.8}                                              \\
\multirow{-6}{*}{\begin{tabular}[c]{@{}c@{}}4.5B\\ (35\%\\Pruned)\end{tabular}} & \multirow{-3}{*}{D\ding{34}} & \cellcolor[HTML]{ECF4FF}Ours: PPL     & \cellcolor[HTML]{ECF4FF}\textbf{23.1} & \multicolumn{1}{c|}{\cellcolor[HTML]{ECF4FF}\textbf{38.8}} & \cellcolor[HTML]{ECF4FF}55.2                                              & \cellcolor[HTML]{ECF4FF}\textbf{1.6}                                     & \cellcolor[HTML]{ECF4FF}\textbf{80.1}                                              & \cellcolor[HTML]{ECF4FF}\textbf{3.4}                                     & \cellcolor[HTML]{ECF4FF}\textbf{37.8}                                              \\ 

\specialrule{.2em}{.1em}{.1em} 

\specialrule{.2em}{.1em}{.1em} 

\multicolumn{3}{c|}{}                                                                                                                               & \multicolumn{3}{c|}{Zero-shot Performance}                                                                                                                                     & \multicolumn{2}{c|}{H100 80GB}                                                                                                                                & \multicolumn{2}{c}{RTX3090 24GB}                                                                                                                              \\ \cline{4-10} 
\multicolumn{3}{c|}{}                                                                                                                               & \multicolumn{2}{c|}{PPL↓}                                                                          &                                                                           &                                                                          &                                                                                    &                                                                          &                                                                                    \\
\multicolumn{3}{c|}{\multirow{-3}{*}{\#Param \& Method}}                                                                                                        & WikiText2                             & \multicolumn{1}{c|}{PTB}                                   & \multirow{-2}{*}{\begin{tabular}[c]{@{}c@{}}Ave Acc↑\\ (\%)\textsuperscript{$\dagger$}
\end{tabular}} & \multirow{-2}{*}{\begin{tabular}[c]{@{}c@{}}Latency↓\\ (s)\end{tabular}} & \multirow{-2}{*}{\begin{tabular}[c]{@{}c@{}}Throughput↑\\ (tokens/s)\end{tabular}} & \multirow{-2}{*}{\begin{tabular}[c]{@{}c@{}}Latency↓\\ (s)\end{tabular}} & \multirow{-2}{*}{\begin{tabular}[c]{@{}c@{}}Throughput↑\\ (tokens/s)\end{tabular}} \\ \hline
\multicolumn{3}{c|}{Vicuna-13B: 13.0B (Original)}                                                                                                        & 14.7                                  & \multicolumn{1}{c|}{51.6}                                  & 68.3                                                                      & 2.8                                                                      & 45.5                                                                               & OOM                                                                      & OOM                                                                                \\ \hline
                                                                                  &                         & Wanda-sp                              & 19.0                                  & \multicolumn{1}{c|}{71.8}                                  & 63.6                                                                      & 3.8                                                                      & 34.1                                                                               & 9.8                                                                      & 12.9                                                                               \\
                                                                                  &                         & FLAP                                  & 18.8                         & \multicolumn{1}{c|}{65.3}                         & 63.3                                                                      & 3.9                                                                      & 32.6                                                                               & 10.2                                                                     & 12.6                                                                               \\
                                                                                  & \multirow{-3}{*}{W\ding{34}} & LLM-Pruner                            & \textbf{16.0}                         & \multicolumn{1}{c|}{57.0}                                  & 65.3                                                                      & 3.8                                                                      & 34.0                                                                               & 7.5                                                                      & 17.3                                                                               \\ \cline{2-10} 
                                                                                  &                         & SLEB                                  & 20.5                                  & \multicolumn{1}{c|}{68.7}                                  & 60.4                                                                      & \textbf{2.3}                                                             & \textbf{55.7}                                                                      & \textbf{5.4}                                                             & \textbf{23.9}                                                                      \\
                                                                                  &                         & \cellcolor[HTML]{ECF4FF}Ours: Taylor+ & \cellcolor[HTML]{ECF4FF}18.1          & \multicolumn{1}{c|}{\cellcolor[HTML]{ECF4FF}61.6}          & \cellcolor[HTML]{ECF4FF}\textbf{66.7}                                     & \cellcolor[HTML]{ECF4FF}\textbf{2.3}                                     & \cellcolor[HTML]{ECF4FF}\textbf{55.7}                                              & \cellcolor[HTML]{ECF4FF}\textbf{5.4}                                     & \cellcolor[HTML]{ECF4FF}\textbf{23.9}                                              \\
\multirow{-6}{*}{\begin{tabular}[c]{@{}c@{}}10.5B\\ (21\%\\Pruned)\end{tabular}} & \multirow{-3}{*}{D\ding{34}} & \cellcolor[HTML]{ECF4FF}Ours: PPL     & \cellcolor[HTML]{ECF4FF}16.1          & \multicolumn{1}{c|}{\cellcolor[HTML]{ECF4FF}\textbf{56.5}} & \cellcolor[HTML]{ECF4FF}64.9                                              & \cellcolor[HTML]{ECF4FF}\textbf{2.3}                                     & \cellcolor[HTML]{ECF4FF}\textbf{55.7}                                              & \cellcolor[HTML]{ECF4FF}\textbf{5.4}                                     & \cellcolor[HTML]{ECF4FF}\textbf{23.9}                                              \\ \hline
\hline
                                                                                  &                         & Wanda-sp                              & 36.6                                  & \multicolumn{1}{c|}{123.5}                                 & 52.7                                                                      & 3.8                                                                      & 33.8                                                                               & 10.5                                                                     & 12.6                                                                               \\
                                                                                  &                         & FLAP                                  & 28.7                                  & \multicolumn{1}{c|}{96.2}                                  & 58.3                                                                      & 3.9                                                                      & 32.9                                                                               & 9.7                                                                      & 13.2                                                                               \\
                                                                                  & \multirow{-3}{*}{W\ding{34}} & LLM-Pruner                            & 22.2                                  & \multicolumn{1}{c|}{74.0}                                  & 59.7                                                            & 3.6                                                                      & 35.6                                                                               & 7.1                                                                      & 18.0                                                                               \\ \cline{2-10} 
                                                                                  &                         & SLEB                                  & 41.6                                  & \multicolumn{1}{c|}{116.5}                                 & 49.4                                                                      & \textbf{1.8}                                                             & \textbf{69.7}                                                                      & \textbf{4.0}                                                             & \textbf{31.7}                                                                      \\
                                                                                  &                         & \cellcolor[HTML]{ECF4FF}Ours: Taylor+ & \cellcolor[HTML]{ECF4FF}34.2          & \multicolumn{1}{c|}{\cellcolor[HTML]{ECF4FF}90.4}          & \cellcolor[HTML]{ECF4FF}\textbf{61.4}                                     & \cellcolor[HTML]{ECF4FF}\textbf{1.8}                                     & \cellcolor[HTML]{ECF4FF}\textbf{69.7}                                              & \cellcolor[HTML]{ECF4FF}\textbf{4.0}                                     & \cellcolor[HTML]{ECF4FF}\textbf{31.7}                                              \\
\multirow{-6}{*}{\begin{tabular}[c]{@{}c@{}}8.3B\\ (37\%\\Pruned)\end{tabular}}  & \multirow{-3}{*}{D\ding{34}} & \cellcolor[HTML]{ECF4FF}Ours: PPL     & \cellcolor[HTML]{ECF4FF}\textbf{22.1} & \multicolumn{1}{c|}{\cellcolor[HTML]{ECF4FF}\textbf{73.6}} & \cellcolor[HTML]{ECF4FF}59.1                                              & \cellcolor[HTML]{ECF4FF}\textbf{1.8}                                     & \cellcolor[HTML]{ECF4FF}\textbf{69.7}                                              & \cellcolor[HTML]{ECF4FF}\textbf{4.0}                                     & \cellcolor[HTML]{ECF4FF}\textbf{31.7}                                              \\ 
\specialrule{.2em}{.1em}{.1em} 
\end{tabular}
\begin{tablenotes}[para,flushleft]
\footnotesize
\textsuperscript{$\dagger$}Average accuracy on seven commonsense reasoning tasks. 
\newline
\textsuperscript{$\ddagger$}Measured with 12 input tokens, 128 output tokens, and a batch size of 1 on a single GPU.
\end{tablenotes}
\end{threeparttable}
\end{adjustbox}
\vspace{-0.05in}

\caption{Results with moderate-level pruning on LLaMA-7B (top) and Vicuna-13B-v1.3 (bottom). Our depth pruning (D\ding{34}) with LoRA retraining achieves similar performance to width pruning (W\ding{34}) methods~\cite{wanda,flap,llmpruner} and outperforms the recent SLEB~\cite{song2024sleb}, while effectively accelerating LLM inference. See Table~\ref{supple_lora_results} for detailed results.} \label{table:lora_results}
\vspace{-0.1in}
\end{table*}

\begin{table*}[t]
\centering

\begin{adjustbox}{max width=0.99\linewidth}
\begin{threeparttable}

\begin{tabular}{cc|cccc|cccc|cccc}
\specialrule{.2em}{.1em}{.1em} 
\multicolumn{2}{c|}{Metric}                                                          & \multicolumn{4}{c|}{PPL↓ on WikiText2}                                                                                                                        & \multicolumn{4}{c|}{Ave Acc↑ (\%)\textsuperscript{$\dagger$}}                                                                                                                            & \multicolumn{4}{c}{Throughput↑ (tokens/s)\textsuperscript{$\ddagger$}}                                                                                                                                                                                                                                                                                                                                                                                                 \\ \hline
\multicolumn{2}{c|}{\#Param after Pruning\textsuperscript{$\star$}}                                           & 5.5B                                  & 3.7B                                  & 2.7B                                  & 1.5B                                  & 5.5B                                  & 3.7B                                  & 2.7B                                  & 1.5B                                  & 5.5B                                                                                                     & 3.7B                                                                                                     & 2.7B                                                                                                      & 1.5B                                                                                                      \\ \hline
\multicolumn{1}{l}{}                        & Wanda-sp                               & 24.4                                  & 364.5                                 & 1370.1                                & 8969.3                                & 58.5                                  & 36.7                                  & 37.0                                  & 35.6                                  & 41.7                                                                                                     & 40.5                                                                                                     & 40.7                                                                                                      & 43.5                                                                                                      \\
\multicolumn{1}{l}{}                        & FLAP                                   & 22.0                                  & 63.1                                  & 589.3                                 & 28727.9                               & 61.4                                  & 47.3                                  & 36.7                                  & 34.5                                  & 40.5                                                                                                     & 41.2                                                                                                     & 41.2                                                                                                      & 42.3                                                                                                      \\
\multicolumn{1}{l}{\multirow{-3}{*}{W\ding{34}}} & LLM-Pruner                             & 19.6                                  & 38.8                                  & 66.4                                  & 202.9                                 & 60.1                                  & 50.1                                  & 44.3                                  & 38.4                                  & 43.2                                                                                                     & 43.4                                                                                                     & 43.9                                                                                                      & 44.8                                                                                                      \\ \hline
                                            & SLEB                                   & 25.1                                  & 110.4                                 & 731.5                                 & 18730.8                               & 55.6                                  & 40.2                                  & 39.1                                  & 37.4                                  & \textbf{66.0}                                                                                            & \textbf{84.0}                                                                                            & \textbf{107.4}                                                                                            & \textbf{182.5}                                                                                            \\
                                            & \cellcolor[HTML]{ECF4FF}Ours, LoRA     & \cellcolor[HTML]{ECF4FF}18.8          & \cellcolor[HTML]{ECF4FF}37.0          & \cellcolor[HTML]{ECF4FF}68.9          & \cellcolor[HTML]{ECF4FF}1002.2        & \cellcolor[HTML]{ECF4FF}60.7          & \cellcolor[HTML]{ECF4FF}47.0          & \cellcolor[HTML]{ECF4FF}40.1          & \cellcolor[HTML]{ECF4FF}37.1          & \cellcolor[HTML]{ECF4FF}                                                                                 & \cellcolor[HTML]{ECF4FF}                                                                                 & \cellcolor[HTML]{ECF4FF}                                                                                  & \cellcolor[HTML]{ECF4FF}                                                                                  \\
                                            & \cellcolor[HTML]{ECF4FF}Ours, CPT      & \cellcolor[HTML]{ECF4FF}\textbf{14.3} & \cellcolor[HTML]{ECF4FF}\textbf{16.0} & \cellcolor[HTML]{ECF4FF}\textbf{17.1} & \cellcolor[HTML]{ECF4FF}\textbf{20.5} & \cellcolor[HTML]{ECF4FF}61.5          & \cellcolor[HTML]{ECF4FF}57.1          & \cellcolor[HTML]{ECF4FF}\textbf{55.0} & \cellcolor[HTML]{ECF4FF}\textbf{49.2} & \cellcolor[HTML]{ECF4FF}                                                                                 & \cellcolor[HTML]{ECF4FF}                                                                                 & \cellcolor[HTML]{ECF4FF}                                                                                  & \cellcolor[HTML]{ECF4FF}                                                                                  \\
\multirow{-4}{*}{D\ding{34}}                     & \cellcolor[HTML]{ECF4FF}Ours, CPT$\Rightarrow$LoRA & \cellcolor[HTML]{ECF4FF}14.8          & \cellcolor[HTML]{ECF4FF}16.5          & \cellcolor[HTML]{ECF4FF}17.8          & \cellcolor[HTML]{ECF4FF}21.1          & \cellcolor[HTML]{ECF4FF}\textbf{63.1} & \cellcolor[HTML]{ECF4FF}\textbf{57.4} & \cellcolor[HTML]{ECF4FF}\textbf{55.0} & \cellcolor[HTML]{ECF4FF}49.0          & \multirow{-3}{*}{\cellcolor[HTML]{ECF4FF}\begin{tabular}[c]{@{}c@{}}\textbf{66.0}\\ \normalsize{(1.2×)}\end{tabular}} & \multirow{-3}{*}{\cellcolor[HTML]{ECF4FF}\begin{tabular}[c]{@{}c@{}}\textbf{84.0}\\ \normalsize{(1.6×)}\end{tabular}} & \multirow{-3}{*}{\cellcolor[HTML]{ECF4FF}\begin{tabular}[c]{@{}c@{}}\textbf{107.4}\\ \normalsize{(2.0×)}\end{tabular}} & \multirow{-3}{*}{\cellcolor[HTML]{ECF4FF}\begin{tabular}[c]{@{}c@{}}\textbf{182.5}\\ \normalsize{(3.4×)}\end{tabular}} \\ \hline
\multicolumn{2}{c|}{Vicuna-7B: 6.7B (Original)}                                                 & \multicolumn{4}{c|}{17.1}                                                                                                                                     & \multicolumn{4}{c|}{65.9}                                                                                                                                     & \multicolumn{4}{c}{53.7}                                                                                                                                                                                                                                                                                                                                                                                                                    \\

\specialrule{.2em}{.1em}{.1em} 
\end{tabular}

\begin{tablenotes}[para,flushleft]
\footnotesize
\textsuperscript{$\star$}The pruning ratios of 20\%, 45\%, 60\%, and 80\% lead to 5.5B, 3.7B, 2.7B, and 1.5B parameters, respectively. The PPL criterion is used to obtain our models.
\newline
\textsuperscript{$\dagger$}Average accuracy on seven commonsense reasoning tasks. 
\newline
\textsuperscript{$\ddagger$}Measured with 12 input tokens, 128 output tokens, and a batch size of 1 on an NVIDIA H100 GPU.
\end{tablenotes}
\end{threeparttable}
\end{adjustbox}

\vspace{-0.05in}
\caption{Effectiveness of CPT under high compression rates on Vicuna-7B-v1.3. CPT is essential to regain the performance of extensively pruned models (e.g., fewer than 3.7B parameters), whereas retraining-free~\cite{flap,song2024sleb} and LoRA-based~\cite{llmpruner} approaches yield unsatisfactory results.} \label{table:cpt_results}
\end{table*}

\begin{table*}[ht]
\centering

\begin{adjustbox}{max width=0.98\linewidth}
\begin{threeparttable}
\begin{tabular}{p{1.8cm}|p{14.8cm}}

\specialrule{.2em}{.1em}{.1em} 
Model       & Generated Text 
\\ \hline
\small{Vicuna-7B}    & 
\footnotesize \uline{AI can create a logo in seconds}! You choose the design elements you like best from pre-made templates, and then the AI generates a unique logo design just for you. This is a great way to get a professional-looking logo without hiring ...
\\ \hline
\small{Wanda-sp}    & 
\footnotesize \uline{AI can create a logo in seconds} now. Inso Ison Im Ison Im I I was able can do do do now do now do now Innowhere you you you you you you you you you you you you you you you you you you you you you you you you you you you ...
\\ \hline
 \small{FLAP}        & 
\footnotesize \uline{AI can create a logo in seconds} The V Just ing A A A A A A A A A, A A A A A S Just a. : S . No An , A A the A The The... The The... A, A A A A. A A A This A The Just A A A A The A A The A A A. and The A All T A A.. A the ...
\\ \hline
 \small{LLM-Pruner}  & 
\footnotesize \uline{AI can create a logo in seconds}: 10 seconds! I've created a logo in minutes using an AI. It has only one image in front for 8 minutes. 4 hours = 65.000 + 12 hours Words: English
Speech: 100 words in 5 minutes 1. 60 hours 2. 492 ...
\\ \hline
\small{SLEB} & 
\footnotesize \uline{AI can create a logo in seconds} while.  :). I have put. I believe it . →.]. Here least →… </<erus known). See →www.giftoings . </<.next]; info. I’ve don-> .…erutex“
Here for. to “Itin.g for the next….. .0.>1260070uro.`s- ...
\\ \hline
\small{Ours, CPT}   & 
\footnotesize \uline{AI can create a logo in seconds}. Even if you don't have a designer who knows the best layouts to use or what colors work best together, AI is already hard at work creating the perfect combination to your artwork.
AI is also capable of ...
\\

\specialrule{.2em}{.1em}{.1em}
\end{tabular}
\end{threeparttable}
\end{adjustbox}

\vspace{-0.05in}
\caption{Generation examples from the original Vicuna-7B and the 60\%-pruned models with 2.7B parameters.}\label{gen_output_2.7b}
\end{table*}

\section{Experimental Setup}

\paragraph{Source Model.} Our testbed includes LLaMA-7B~\cite{touvron2023llama} and Vicuna-\{7B, 13B\}-v1.3~\cite{vicuna}, which are famous LLMs.

\paragraph{Baseline.} LLM-Pruner~\cite{llmpruner}, FLAP~\cite{flap}, and Wanda-sp (i.e., a structured variant~\cite{flap} of Wanda~\cite{wanda}) serve as the baselines for width pruning. Table~\ref{table:arch_short_ver} shows the pruned architectures under similar numbers of parameters. We also examine SLEB~\cite{song2024sleb}, a retraining-free block pruning method for LLMs, which has been concurrently introduced with our study. Section~\ref{sec:supple_baseline} describes the baselines in detail.

\paragraph{Data.} Following~\citet{llmpruner}, we randomly select 10 samples from BookCorpus~\cite{Zhu_2015_ICCV} to compute block-level significance during the pruning stage. We also use this calibration dataset for the baseline methods to ensure a fair comparison. In LoRA retraining, 50K samples of the refined Alpaca~\cite{alpaca} are used for instruction tuning. In CPT retraining, we leverage SlimPajama~\cite{cerebras2023slimpajama}, which consists of 627B tokens for LLM pretraining. 

\paragraph{Evaluation.} Following~\citet{touvron2023llama}, we measure zero-shot accuracy on commonsense reasoning datasets (i.e., BoolQ~\cite{clark-etal-2019-boolq}, PIQA~\cite{Bisk2020piqa}, HellaSwag~\cite{zellers2019hellaswag}, WinoGrande~\cite{sakaguchi2019winogrande}, ARC-easy~\cite{clark2018think}, ARC-challenge~\cite{clark2018think}, and OpenbookQA~\cite{OpenBookQA2018}) using the lm-evaluation-harness package~\cite{eval-harness}. We also report zero-shot PPL on WikiText2~\cite{wikitext2} and PTB~\cite{marcus-etal-1993-building}.

\paragraph{Latency and Throughput.} We follow~\citet{sheng2023flexgen} to measure the metrics. Given a batch size $M$ and an output sequence length $L$ (excluding the input length), the latency $T$ represents the time required to handle the given prompts and produce $ML$ output tokens. The throughput is computed as $ML/T$. We report the average results from 20 runs after the initial 10 warm-up batches.

\paragraph{Implementation.} We use the Hugging Face's Transformers library~\cite{wolf-etal-2020-transformers}. For pruning and LoRA retraining, an NVIDIA A100 GPU is employed. For CPT retraining, eight NVIDIA H100 GPUs are utilized, with a training duration of less than two weeks for each model size. For inference, we opt for the default setup of the Transformers library. See Section~\ref{sec:supple_impl_details} for the details.

\section{Results}

\subsection{Moderate Pruning and LoRA Retraining}
Tables~\ref{table:lora_results}~and~\ref{supple_lora_results} show the zero-shot performance and inference efficiency of differently pruned models. Here, our models are obtained using a light LoRA retraining setup. The width pruning methods~\cite{llmpruner,flap,wanda} do not improve LLM inference efficiency. Under limited input (batch) scales, the processing speed largely hinges on the frequency of memory access operations. Addressing this issue by merely reducing matrix sizes is challenging, unless they are completely removed. The speed even worsens compared to the original model due to GPU-unfriendly operation dimensions (e.g., the hidden sizes of FFN are often not divisible by 8 (Table~\ref{supple_table:arch}), which hinders the effective utilization of GPU Tensor Cores~\cite{tensor_core_guide}). 

On the contrary, our depth pruning exhibits speedups through the complete removal of several Transformer blocks, resulting in fewer memory access and matrix-level operations between activations and weights. Moreover, under the same LoRA retraining protocol as~\citet{llmpruner}, our models achieve zero-shot scores on par with finely width-pruned models. Although SLEB~\cite{song2024sleb} enhances inference efficiency similar to our method, its approach without retraining falls short in developing proficient small LLMs. See Section~\ref{sec:supple_moderate_pruning} for detailed results.

\begin{figure}[t]
  \centering
    \includegraphics[width=\linewidth]{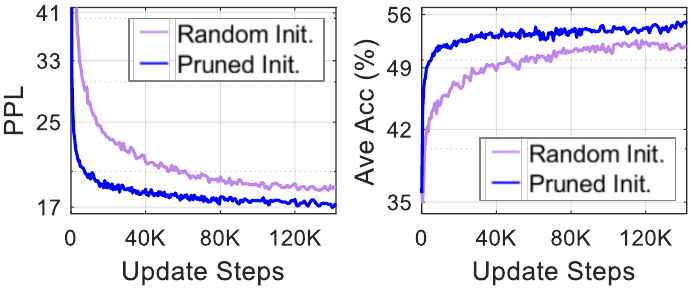}
  \caption{Zero-shot scores during the training progress of the 2.7B-parameter model from Vicuna-7B. Using the pruned network as initialization (blue lines) for CPT accelerates the learning process and yields better results than starting from scratch (purple).}
  \label{fig_learncurve_cpt}
  \vspace{-0.2in}
\end{figure}

\subsection{Aggressive Pruning and CPT Retraining}
Table~\ref{table:cpt_results} compares different retraining methods. Our models are obtained using the PPL criterion. Under high pruning ratios (e.g., yielding fewer than 3.7B parameters), LoRA-based tuning (LLM-Pruner~\cite{llmpruner}; Ours, LoRA) and retraining-free approaches (Wanda-sp~\cite{wanda,flap}, FLAP~\cite{flap}, SLEB~\cite{song2024sleb}) fail to recover model performance. In contrast, CPT proves effective in regaining the quality of heavily pruned models. CPT$\Rightarrow$LoRA slightly improves zero-shot accuracy for some pruning ratios, but with a minor drop in PPL. Table~\ref{gen_output_2.7b} presents samples produced by 2.7B-parameter models (60\% pruned). In contrast to the baselines, our model can generate text that is fluent and appropriately aligned with the context.

Compared to LoRA retraining, the computational costs for CPT are considerably higher: LoRA can be completed within a day using just one GPU, while CPT requires about two weeks with eight GPUs in our experiments, with the option to use more if needed. However, utilizing a pruned network for initialization in CPT leads to faster learning and better results than building the same-sized models from scratch (see Figure~\ref{fig_learncurve_cpt}), highlighting its efficacy for smaller LLMs. Section \ref{sec:supple_aggressive_pruning} presents the learning progress in detail.

\begin{figure}[t]
  \centering
    \includegraphics[width=\linewidth]{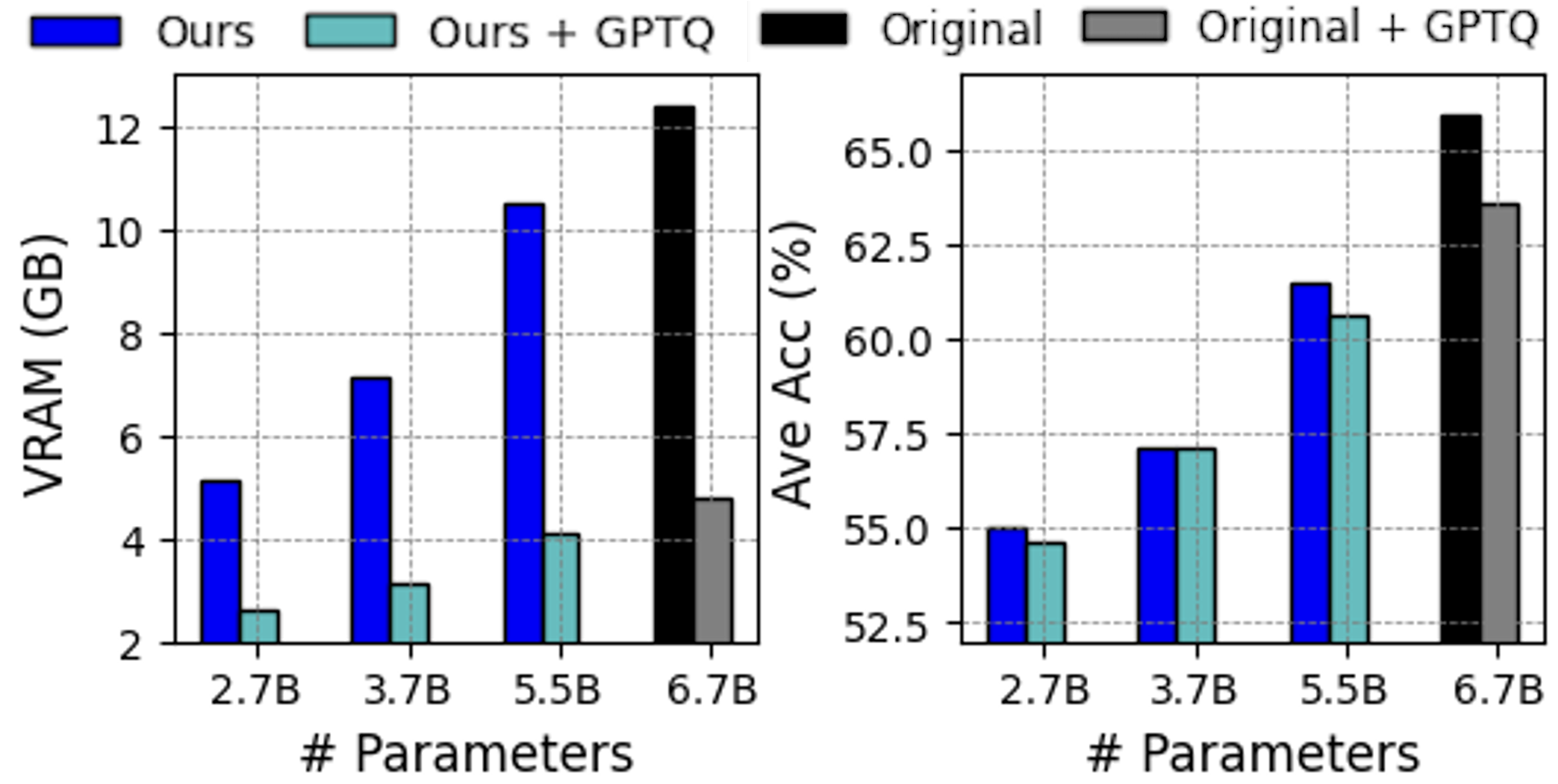}
        \vspace{-0.25in}
  \caption{Further compression with GPTQ. Our pruned models following 4-bit weight quantization exhibit reduced VRAM usage without significant performance decline. The results for the original Vicuna-7B are presented for reference. See Section~\ref{sec:appendix_gptq} for the details.}
  \vspace{-0.1in}
  \label{fig_gptq_measure}
\end{figure}

\subsection{Applicability with Quantization} 
Leveraging post-training quantization (PTQ) effectively lowers the memory consumption for inference of LLMs. Figure~\ref{fig_gptq_measure} shows the results of applying GPTQ~\cite{frantar2023optq}, a well-known PTQ method, to our depth-pruned models after CPT. The 4-bit weight quantization significantly reduces the VRAM demands across various model sizes without noticeable degradation in zero-shot accuracy. See Section~\ref{sec:appendix_gptq} for further results.

\subsection{Ablation Study}
We explore various design factors, including the criteria for importance evaluation, the choice of units for depth pruning, and the impact of calibration data volume. The results presented in this section were obtained through LoRA retraining.

\subsubsection{Importance Criteria for Block Pruning} 
Table~\ref{table:criterion} presents the results of block pruning using various significance criteria. The basic methods without the `+' label fail to maintain essential initial blocks, causing a decline in performance. The Mag+ method, which preserves these critical blocks, partially improves the scores; however, its effectiveness is still inferior compared to the other methods, indicating that relying solely on weight magnitude could be improper for pruning decisions. The Taylor+ criterion enhances accuracy in commonsense reasoning tasks, while the PPL method leads to better generation quality without relying on heuristic selection of pruning candidates.

\subsubsection{Structural Unit for Depth Pruning} 
Pruning individual MHA and FFN modules, which are more fine-grained units than Transformer blocks, is also possible. To examine its effect, we measure the impact of removing each module on the PPL of the calibration set and selectively eliminate the unnecessary modules. The same LoRA retraining procedure is conducted.

\begin{table}[t]
\centering

\begin{adjustbox}{max width=0.88\columnwidth}
\begin{threeparttable}
\begin{tabular}{cc|cc|c}

\specialrule{.2em}{.1em}{.1em} 

\multicolumn{2}{c|}{\multirow{2}{*}{\begin{tabular}[c]{@{}c@{}}Block Pruning\\ Criterion\end{tabular}}} & \multicolumn{2}{c|}{PPL↓} & \multirow{2}{*}{\begin{tabular}[c]{@{}c@{}}Ave Acc↑\\  (\%)\textsuperscript{$\dagger$} \end{tabular}} \\
\multicolumn{2}{c|}{}                                                                                   & WikiText2    & PTB        &                                                                          \\ \hline
\multirow{5}{*}{\begin{tabular}[c]{@{}c@{}}5.5B\\ (20\%\\Pruned)\end{tabular}}       & Mag       & 7720.7       & 10618.7    & 34.4                                                                     \\
                                                                                      & Mag+      & 19.4         & 36.3       & 56.1                                                                     \\
                                                                                      & Taylor        & 3631.7       & 4327.9     & 35.5                                                                     \\
                                                                                      & Taylor+       & 20.2         & 32.3       & \textbf{63.5}                                                                     \\
                                                                                      & PPL             & \textbf{17.7}         & \textbf{30.7}       & 61.9                                                                     \\ \hline
\multirow{5}{*}{\begin{tabular}[c]{@{}c@{}}4.5B\\ (35\%\\Pruned)\end{tabular}}       & Mag       & 8490.1       & 14472.1    & 34.9                                                                     \\
                                                                                      & Mag+      & 36.9         & 61.1       & 49.3                                                                     \\
                                                                                      & Taylor        & 7666.8       & 10913.1    & 35.3                                                                     \\
                                                                                      & Taylor+       & 33.2         & 58.5       & \textbf{55.4}                                                                     \\
                                                                                      & PPL            & \textbf{23.1}         & \textbf{38.8}       & 55.2                                                                     \\ 

\specialrule{.2em}{.1em}{.1em} 
\end{tabular}
\begin{tablenotes}[para,flushleft]
\footnotesize
\textsuperscript{$\dagger$}Average accuracy on seven commonsense reasoning tasks. 
\end{tablenotes}
\end{threeparttable}
\end{adjustbox}

\vspace{-0.05in}
  
\caption{Comparison of pruning criteria on LLaMA-7B. The Taylor+ method excels in commonsense reasoning accuracy, while the PPL criterion leads to better generation performance.} \label{table:criterion}

\vspace{+0.15in}

\begin{adjustbox}{max width=\columnwidth}
\begin{threeparttable}
\begin{tabular}{c|c|cc|c}

\specialrule{.2em}{.1em}{.1em} 

\multirow{2}{*}{\begin{tabular}[c]{@{}c@{}}Depth Pruning\\ Unit\end{tabular}} & \multirow{2}{*}{\#Param} & \multicolumn{2}{c|}{PPL↓} & \multirow{2}{*}{\begin{tabular}[c]{@{}c@{}}Ave Acc↑\\  (\%)\textsuperscript{$\dagger$} \end{tabular}} \\
                                                                              &                          & WikiText2      & PTB      &                                                                          \\ \hline
Individual MHA \& FFN                                                                & 5.7B                     & 20.8           & 34.8     & \textbf{63.1}                                                                     \\
Transformer Block                                                             & 5.7B                     & \textbf{16.9}           & \textbf{29.3}     & 62.8                                                                     \\ \hline
Individual MHA \& FFN                                                                & 5.3B                     & 25.2           & 41.3     & \textbf{61.1}                                                                     \\
Transformer Block                                                             & 5.3B                     & \textbf{18.6}           & \textbf{33.1}     & 60.6                                                                     \\ \hline
Individual MHA \& FFN                                                                & 4.6B                     & 38.9           & 58.7     & 52.5                                                                     \\
Transformer Block                                                             & 4.5B                     & \textbf{23.1}           & \textbf{38.8}     & \textbf{55.2}                                                                     \\ \hline
Individual MHA \& FFN                                                                & 4.0B                     & 63.2           & 88.9     & 48.3                                                                     \\
Transformer Block                                                             & 3.9B                     & \textbf{31.1}           & \textbf{47.3}     & \textbf{50.6}                                                                     \\ 

\specialrule{.2em}{.1em}{.1em} 
\end{tabular}
\begin{tablenotes}[para,flushleft]
\footnotesize
\textsuperscript{$\dagger$}Average accuracy on seven commonsense reasoning tasks. 
\end{tablenotes}
\end{threeparttable}
\end{adjustbox}

\vspace{-0.05in}
\caption{Comparison of depth pruning granularities on LLaMA-7B. Removing entire Transformer blocks instead of individual MHA and FFN modules generally yields better results.} \label{table:ablation_pruneunit}
\vspace{-0.1in}
\end{table}

Table~\ref{table:ablation_pruneunit} shows the results of depth pruning at different granularities. For the models with more than 5B parameters, removing individual MHA and FFN modules results in better downstream task accuracy but worse PPL compared to removing entire Transformer blocks. For smaller models than 5B, block-level pruning achieves superior results in terms of all the examined metrics. This differs from the common belief that removing finer units yields better performance.

Given the collaborative roles of the modules (i.e., MHA captures dependency relations~\cite{transformer}, while skip connections and FFN prevent the rank collapse in purely attention-driven networks~\cite{dong2021attention}), it may be suboptimal to treat them in isolation. Taking the 5.3B model in Table~\ref{table:ablation_pruneunit} as an example, module-level pruning results in consecutive FFNs in some positions, potentially impairing the model's ability to handle word interactions. In contrast, with block-level removal, the loss of information could be compensated by neighboring blocks that serve similar functions.

\subsubsection{Calibration Data Volume}
 The calibration set is employed to assess the weight significance of width pruning baselines and the block-level importance of our method during the pruning phase. Table~\ref{table:calib_data_volume} presents the results obtained by varying the number of calibration samples in the BookCorpus dataset. The scores remain relatively stable for the examined methods, suggesting that 10 samples could be sufficient. However, our Taylor+ method encounters a drop in downstream task accuracy when 1K samples are used, leaving the exploration of calibration data characteristics for future research.

\section{Related Work} \label{relwork}

Numerous techniques have been developed towards efficient LLMs, including knowledge distillation~\cite{fu2023specializing,hsieh2023distilling}, quantization~\cite{frantar2023optq,dettmers2022llmint8}, and system-level inference acceleration~\cite{dao2023flashattention2,kwon2023efficient}. In this study, we focus on network pruning~\cite{lecun1989optimal}, which has a long-standing reputation in the model compression field. Beyond its use in relatively small-scale convolutional networks~\cite{li2016pruning,he2019filter} and Transformer models~\cite{yu2022unified,xia2022structured,kurtic2023ziplm}, pruning has recently begun to be applied to contemporary LLMs. Several studies~\cite{frantar-sparsegpt,wanda} employ unstructured and semi-structured~\cite{zhou2021} pruning by zeroing individual neurons. SparseGPT~\cite{frantar-sparsegpt} addresses the layer-wise reconstruction problem for pruning by computing Hessian inverses. Wanda~\cite{wanda} introduces a pruning criterion that involves multiplying weight magnitudes by input feature norms. Despite the plausible performance of pruned models using zero masks, they necessitate specialized support for sparse matrix operations to ensure actual speedups.

In contrast, structured pruning removes organized patterns, such as layers~\cite{fan2019reducing,jha2023train}, MHA's attention heads~\cite{voita2019analyzing,michel2019sixteen}, FFN's hidden sizes~\cite{nova2023gradientfree,santacroce2023matters}, and some hybrid forms~\cite{lagunas2021block,xia2022structured,kwon2022fast,kurtic2023ziplm}, thereby improving inference efficiency in a hardware-agnostic way. To compress LLMs, FLAP~\cite{flap} and LLM-Pruner~\cite{llmpruner} eliminate coupled structures in the aspect of network width while retaining the number of layers. Sheared-LLaMA~\cite{xia2023sheared} introduces a mask learning phase aimed at identifying prunable components in both the network's width and depth. Our study explores the relatively untapped area of depth-only pruning for multi-billion parameter LLMs, which can markedly accelerate latency while attaining competitive performance.

Strategies for skipping layers~\cite{schuster2022confident,delcorro2023skipdecode,raposo2024mixtureofdepths} effectively serve to decrease computational burdens. Moreover, depth pruning approaches~\cite{song2024sleb,men2024shortgpt,tang2024rethinking} for LLMs have been proposed concurrently with our work, based on the architectural redundancy in LLMs.

\begin{table}[t]
\centering
\begin{adjustbox}{max width=0.94\columnwidth}
\begin{threeparttable}
\begin{tabular}{cc|cccc}

\specialrule{.2em}{.1em}{.1em} 

\multirow{2}{*}{\begin{tabular}[c]{@{}c@{}}Evaluation\\ Metric\end{tabular}} & \multirow{2}{*}{Method} & \multicolumn{4}{c}{\# Calibration Samples}                    \\
                                                                             &                         & 10            & 50            & 100           & 1000          \\ \hline
\multirow{5}{*}{\begin{tabular}[c]{@{}c@{}}PPL↓ on\\ WikiText2\end{tabular}} & Wanda-sp                & 21.4          & 21.4          & 21.7          & 20.8          \\
                                                                             & FLAP                    & \textbf{17.0} & 17.5          & 17.5          & \textbf{17.3} \\
                                                                             & LLM-Pruner              & 17.6          & \textbf{17.2} & \textbf{17.0} & OOM\textsuperscript{$\ddagger$}           \\
                                                                             & Ours: Taylor+             & 20.2          & 20.2          & 19.0          & 19.6          \\
                                                                             & Ours: PPL               & 17.7          & \textbf{17.2} & 17.4          & 17.4          \\ \hline
\multirow{5}{*}{\begin{tabular}[c]{@{}c@{}}Ave Acc↑\\  (\%)\textsuperscript{$\dagger$} \end{tabular}}     & Wanda-sp                & 51.8          & 52.9          & 52.0          & 53.0          \\
                                                                             & FLAP                    & 59.5          & 59.7          & 59.9          & 60.8          \\
                                                                             & LLM-Pruner              & 61.8          & 61.6          & 61.7          & OOM\textsuperscript{$\ddagger$}           \\
                                                                             & Ours: Taylor+             & \textbf{63.5} & \textbf{63.5} & \textbf{63.9} & \textbf{61.7} \\
                                                                             & Ours: PPL               & 61.9          & 61.5          & 61.7          & \textbf{61.7} \\

\specialrule{.2em}{.1em}{.1em} 

\end{tabular}
\begin{tablenotes}[para,flushleft]
\footnotesize
\textsuperscript{$\dagger$}Average accuracy on seven commonsense reasoning tasks. 
\newline
\textsuperscript{$\ddagger$}Out-of-memory error on an A100 (80GB) using the official code. 
\end{tablenotes}
\end{threeparttable}
\end{adjustbox}

\vspace{-0.05in} 
\caption{Impact of calibration data volume. The results of 20\%-pruned LLaMA-7B are reported.} \label{table:calib_data_volume}
\vspace{-0.1in}
\end{table}

\section{Conclusion}
By introducing a block pruning method, we conduct an in-depth comparative analysis on the impact of network width and depth on LLM compression. Our work involves the one-shot removal of Transformer blocks. Despite its simplicity, our method with light LoRA retraining matches the zero-shot capabilities of recent width pruning techniques under moderate pruning levels. Moreover, it offers significant inference speedups in resource-constrained scenarios that require running LLMs with limited batch sizes, where width pruning falls short. When comparing retraining strategies, continued pretraining on a large-scale dataset significantly surpasses LoRA-based tuning, particularly in cases of severe pruning. We hope this study will support the development of potent small LLMs.

\section*{Limitations}
Due to constraints in computational resources, we could not test our method on LLMs exceeding 13B parameters. We plan to explore larger models in future research, given that our method can be applied to any model size. Secondly, we found that continued pretraining was essential for performance recovery after extensive pruning. Further exploration of different training corpora and hyperparameters could lead to additional performance improvements. Lastly, commercially available LLMs are optimized for human preferences, such as safety and helpfulness, through alignment tuning. We have yet to assess human preferences throughout the entire process of pruning, retraining, and quantization. We hope future research will address this aspect.

\section*{Acknowledgments}
We thank the Microsoft Startups Founders Hub program and the AI Industrial Convergence Cluster Development project funded by the Ministry of Science and ICT (MSIT, Korea) and Gwangju Metropolitan City for their generous support of GPU resources.

\bibliography{custom}
\clearpage
\appendix
\onecolumn
\begin{center}
{\bf {\Large Appendix — Shortened LLaMA: Depth Pruning for LLMs}} 
\newline
\end{center}

\vspace{-0.15in} 
\section{Additional Results of Inference Efficiency} 

\subsection{Latency-Throughput Trade-Off} \label{sec:appendix_efficiency}
\begin{figure}[h]
  \centering
    \includegraphics[width=\linewidth]{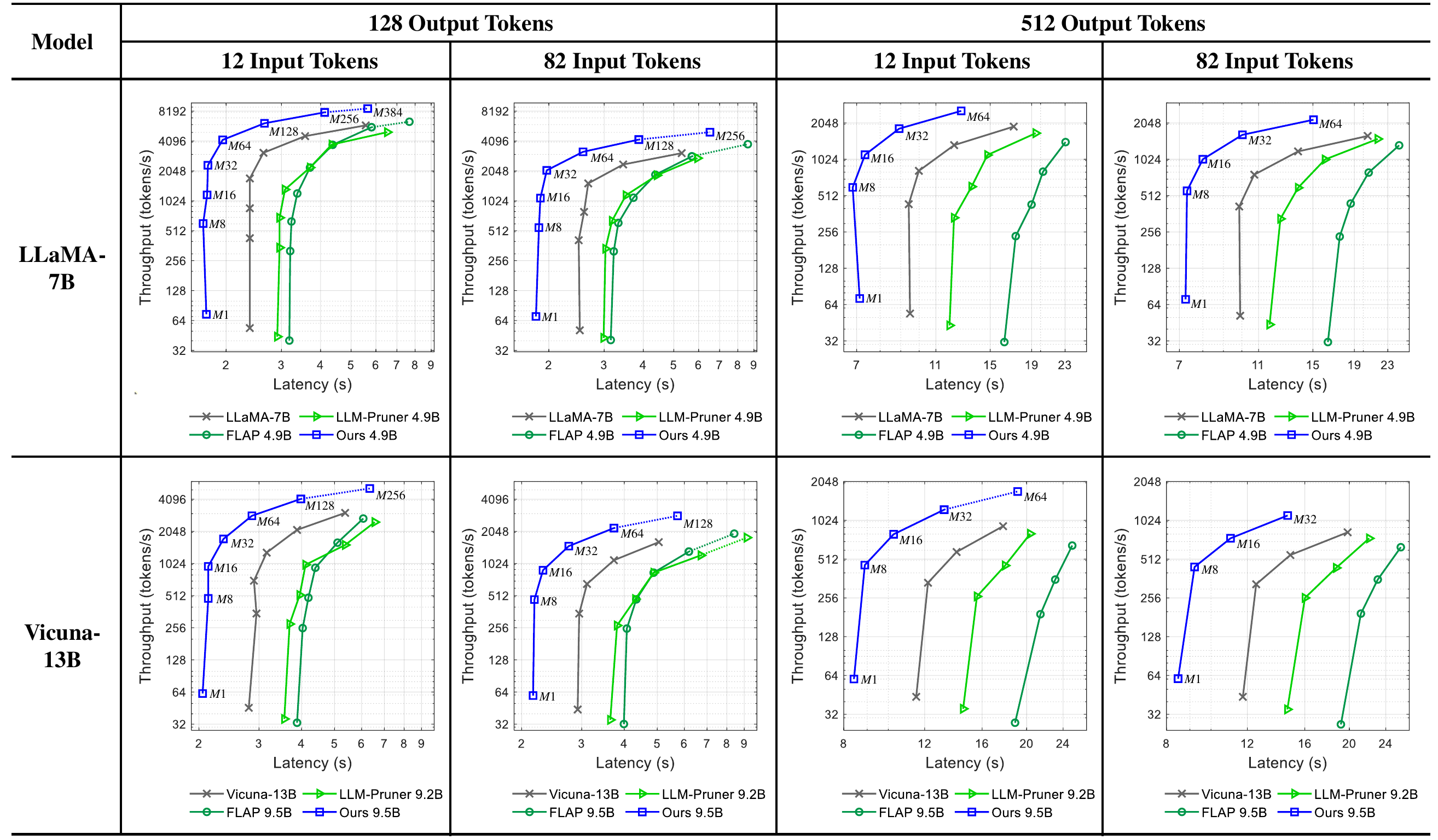}
        \vspace{-0.2in}
  \caption{Inference efficiency of pruned models on an NVIDIA H100 GPU.}  
  \label{fig_appendix_thput_lat}
\end{figure}

As shown in Figure~\ref{fig_appendix_thput_lat}, our depth pruning achieves a superior latency-throughput trade-off for various sequence lengths of input and output. In contrast, the width pruning of FLAP~\cite{flap} and LLM-Pruner~\cite{llmpruner} degrades efficiency results due to GPU-unfriendly weight dimensions~\cite{tensor_core_guide} (e.g., the hidden sizes of FFN are often not divisible by 8). The markers labeled with $M$ represent batch sizes. The dotted lines indicate that pruned models can operate with larger batch sizes, avoiding out-of-memory errors encountered by the original model.

\vspace{-0.05in} 
\subsection{GPU Memory Requirements} 
Table~\ref{table:supple_memory} shows the gains in VRAM usage from our pruned models on an NVIDIA H100 given 12 input tokens. The larger the batch size, the greater the improvement observed. Notably, our pruned models can handle an output length of 512 and a batch size of 64, unlike the original 13B-parameter model.

\begin{table}[h]
\centering
\begin{adjustbox}{max width=0.72\columnwidth}
\begin{threeparttable}
\begin{tabular}{c|ccc|ccc}

\specialrule{.2em}{.1em}{.1em} 
\multirow{2}{*}{\#Param} & \multicolumn{3}{c|}{$L$128} & \multicolumn{3}{c}{$L$512} \\
                       & $M$1    & $M$16   & $M$64   & $M$1    & $M$16   & $M$64  \\ \hline
6.7B (Original)            & 12.8GB  & 16.0GB  & 25.8GB  & 13.3GB  & 25.0GB  & 61.8GB \\ \hline
5.5B (20\% Pruned)             & 10.5GB  & 13.1GB  & 21.1GB  & 10.9GB  & 20.4GB  & 50.4GB \\
4.9B (27\% Pruned)             & 9.4GB   & 11.6GB  & 18.8GB  & 9.7GB   & 18.1GB  & 44.6GB \\
4.5B (35\% Pruned)             & 8.6GB   & 10.7GB  & 17.2GB  & 9.0GB   & 16.6GB  & 40.8GB \\ 
\specialrule{.2em}{.1em}{.1em} 
13.0B (Original)           & 24.8GB  & 29.6GB  & 44.9GB  & 25.5GB  & 43.7GB  & OOM    \\ \hline
10.5B (21\% Pruned)            & 19.9GB  & 23.8GB  & 36.0GB  & 20.5GB  & 35.0GB  & OOM    \\
9.5B (29\% Pruned)            & 18.1GB  & 21.7GB  & 32.7GB  & 18.6GB  & 31.8GB  & 73.5GB \\
8.3B (37\% Pruned)            & 15.7GB  & 18.8GB  & 28.3GB  & 16.1GB  & 27.5GB  & 63.5GB \\ 
\specialrule{.2em}{.1em}{.1em} 
\end{tabular}
\end{threeparttable}
\end{adjustbox}

\vspace{-0.05in}
\caption{GPU memory requirements for varying sequence lengths ($L$) and batch sizes ($M$). The results of the 7B and 13B models and our models with different pruning ratios are reported. Our approach effectively reduces the memory demands of the original models.} \label{table:supple_memory}
\vspace{-0.1in}
\end{table}

\clearpage
\section{Further Results of Moderate Pruning and LoRA Retraining}\label{sec:supple_moderate_pruning}

\subsection{Zero-shot Downstream Task Performance}\label{sec:supple_zeroshot}
\vspace{-0.08in}  
\begin{table*}[ht!]
\centering
\vspace{-0.04in}
\begin{adjustbox}{max width=0.994\linewidth}
\begin{threeparttable}
\begin{tabular}{cc|cc|cccccccc|cc}

\specialrule{.2em}{.1em}{.1em}

\multicolumn{2}{c|}{}                                                                                                  & \multicolumn{2}{c|}{PPL↓}                                                     & \multicolumn{8}{c|}{Commonsense Reasoning Accuracy↑ (\%)}                                                                                                                                                                                                                                                                                 & \multicolumn{2}{c}{Thr↑ (tokens/s)\textsuperscript{$\ddagger$}}                                           \\
\multicolumn{2}{c|}{\multirow{-2}{*}{\#Param \& Method}}                                                                           & Wiki2                                 & PTB                                   & \multicolumn{1}{c|}{Average}                               & BoolQ                                 & PIQA                         & HellaSwag                             & WinoGrande                            & ARC-e                                 & ARC-c                                 & OBQA                                  & H100                                  & RTX3090                               \\ \hline
\multicolumn{2}{c|}{LLaMA-7B: 6.7B}                                                                                    & 12.6                                  & 22.1                                  & \multicolumn{1}{c|}{66.3}                                  & 75.0                                  & 78.7                         & 76.2                                  & 69.9                                  & 75.3                                  & 44.7                                  & 44.4                                  & 53.7                                  & 25.0                                  \\ \hline
                                                                                 & Wanda-sp                            & 21.4                                  & 47.2                                  & \multicolumn{1}{c|}{51.8}                                  & 61.5                                  & 70.4                         & 53.2                                  & 56.0                                  & 58.7                                  & 31.4                                  & 31.0                                  & 41.7                                  & 16.7                                  \\
                                                                                 & FLAP                                & \textbf{17.0}                         & \textbf{30.1}                         & \multicolumn{1}{c|}{59.5}                                  & 69.4                                  & 74.7                         & 66.9                                  & 66.3                                  & 64.6                                  & 36.5                                  & 38.2                                  & 40.5                                  & 16.5                                  \\
                                                                                 & LLM-Pruner                          & 17.6                                  & 30.4                                  & \multicolumn{1}{c|}{61.8}                                  & 66.2                                  & \textbf{77.6}                & 71.4                                  & 66.1                                  & \textbf{70.5}                         & 39.3                                  & \textbf{41.2}                         & 43.2                                  & 21.4                                  \\
                                                                                 & SLEB                                & 18.5                                  & 31.6                                  & \multicolumn{1}{c|}{57.6}                                  & 65.0                                  & 75.0                         & 65.7                                  & 57.9                                  & 67.6                                  & 36.6                                  & 35.8                                  & \textbf{66.0}                         & \textbf{28.4}                         \\
                                                                                 & \cellcolor[HTML]{ECF4FF}Ours: Grad+ & \cellcolor[HTML]{ECF4FF}20.2          & \cellcolor[HTML]{ECF4FF}32.3          & \multicolumn{1}{c|}{\cellcolor[HTML]{ECF4FF}\textbf{63.5}} & \cellcolor[HTML]{ECF4FF}\textbf{75.7} & \cellcolor[HTML]{ECF4FF}75.7 & \cellcolor[HTML]{ECF4FF}\textbf{71.5} & \cellcolor[HTML]{ECF4FF}\textbf{69.1} & \cellcolor[HTML]{ECF4FF}69.9          & \cellcolor[HTML]{ECF4FF}\textbf{41.6} & \cellcolor[HTML]{ECF4FF}40.8          & \cellcolor[HTML]{ECF4FF}\textbf{66.0} & \cellcolor[HTML]{ECF4FF}\textbf{28.4} \\
\multirow{-6}{*}{\begin{tabular}[c]{@{}c@{}}5.5B\\ (20\%\\ Pruned)\end{tabular}} & \cellcolor[HTML]{ECF4FF}Ours: PPL   & \cellcolor[HTML]{ECF4FF}17.7          & \cellcolor[HTML]{ECF4FF}30.7          & \multicolumn{1}{c|}{\cellcolor[HTML]{ECF4FF}61.9}          & \cellcolor[HTML]{ECF4FF}72.7          & \cellcolor[HTML]{ECF4FF}75.7 & \cellcolor[HTML]{ECF4FF}70.4          & \cellcolor[HTML]{ECF4FF}63.6          & \cellcolor[HTML]{ECF4FF}69.5          & \cellcolor[HTML]{ECF4FF}40.1          & \cellcolor[HTML]{ECF4FF}\textbf{41.2} & \cellcolor[HTML]{ECF4FF}\textbf{66.0} & \cellcolor[HTML]{ECF4FF}\textbf{28.4} \\ \hline
                                                                                 & Wanda-sp                            & 50.4                                  & 106.9                                 & \multicolumn{1}{c|}{42.1}                                  & 62.0                                  & 60.4                         & 33.2                                  & 52.8                                  & 37.6                                  & 23.0                                  & 25.4                                  & 41.7                                  & 16.0                                  \\
                                                                                 & FLAP                                & 21.3                                  & 37.1                                  & \multicolumn{1}{c|}{55.8}                                  & 68.2                                  & 70.6                         & 61.0                                  & 64.1                                  & 58.8                                  & 31.4                                  & 36.8                                  & 40.2                                  & 16.5                                  \\
                                                                                 & LLM-Pruner                          & \textbf{20.5}                         & 36.1                                  & \multicolumn{1}{c|}{58.7}                                  & 62.8                                  & \textbf{75.5}                & \textbf{67.2}                         & 64.9                                  & 63.5                                  & 36.8                                  & \textbf{40.2}                         & 44.0                                  & 22.9                                  \\
                                                                                 & SLEB                                & 25.3                                  & 41.3                                  & \multicolumn{1}{c|}{52.6}                                  & 62.1                                  & 71.1                         & 57.2                                  & 53.3                                  & 57.5                                  & 31.6                                  & 35.6                                  & \textbf{73.9}                         & \textbf{34.9}                         \\
                                                                                 & \cellcolor[HTML]{ECF4FF}Ours: Grad+ & \cellcolor[HTML]{ECF4FF}29.9          & \cellcolor[HTML]{ECF4FF}42.0          & \multicolumn{1}{c|}{\cellcolor[HTML]{ECF4FF}\textbf{59.8}} & \cellcolor[HTML]{ECF4FF}\textbf{70.6} & \cellcolor[HTML]{ECF4FF}73.0 & \cellcolor[HTML]{ECF4FF}65.7          & \cellcolor[HTML]{ECF4FF}\textbf{68.5} & \cellcolor[HTML]{ECF4FF}63.9          & \cellcolor[HTML]{ECF4FF}\textbf{39.3} & \cellcolor[HTML]{ECF4FF}37.4          & \cellcolor[HTML]{ECF4FF}\textbf{73.9} & \cellcolor[HTML]{ECF4FF}\textbf{34.9} \\
\multirow{-6}{*}{\begin{tabular}[c]{@{}c@{}}4.9B\\ (27\%\\ Pruned)\end{tabular}} & \cellcolor[HTML]{ECF4FF}Ours: PPL   & \cellcolor[HTML]{ECF4FF}20.7          & \cellcolor[HTML]{ECF4FF}\textbf{36.0} & \multicolumn{1}{c|}{\cellcolor[HTML]{ECF4FF}57.6}          & \cellcolor[HTML]{ECF4FF}66.6          & \cellcolor[HTML]{ECF4FF}73.1 & \cellcolor[HTML]{ECF4FF}63.7          & \cellcolor[HTML]{ECF4FF}60.4          & \cellcolor[HTML]{ECF4FF}\textbf{64.3} & \cellcolor[HTML]{ECF4FF}36.0          & \cellcolor[HTML]{ECF4FF}39.2          & \cellcolor[HTML]{ECF4FF}\textbf{73.9} & \cellcolor[HTML]{ECF4FF}\textbf{34.9} \\ \hline
                                                                                 & Wanda-sp                            & 133.6                                 & 210.1                                 & \multicolumn{1}{c|}{36.9}                                  & 44.5                                  & 56.8                         & 29.6                                  & 49.6                                  & 31.7                                  & 20.7                                  & 25.6                                  & 41.6                                  & 16.1                                  \\
                                                                                 & FLAP                                & 25.6                                  & 44.4                                  & \multicolumn{1}{c|}{52.7}                                  & \textbf{68.3}                         & 68.1                         & 55.9                                  & 61.1                                  & 52.3                                  & 29.4                                  & 33.8                                  & 40.5                                  & 15.8                                  \\
                                                                                 & LLM-Pruner                          & 24.2                                  & 40.7                                  & \multicolumn{1}{c|}{\textbf{55.5}}                         & 62.9                                  & \textbf{72.8}                & \textbf{62.3}                         & 62.7                                  & 57.4                                  & 33.0                                  & \textbf{37.6}                         & 44.4                                  & 21.1                                  \\
                                                                                 & SLEB                                & 34.2                                  & 49.8                                  & \multicolumn{1}{c|}{50.1}                                  & 62.2                                  & 69.0                         & 52.7                                  & 52.9                                  & 51.6                                  & 29.9                                  & 32.2                                  & \textbf{80.1}                         & \textbf{37.8}                         \\
                                                                                 & \cellcolor[HTML]{ECF4FF}Ours: Grad+ & \cellcolor[HTML]{ECF4FF}33.2          & \cellcolor[HTML]{ECF4FF}58.5          & \multicolumn{1}{c|}{\cellcolor[HTML]{ECF4FF}55.4}          & \cellcolor[HTML]{ECF4FF}62.5          & \cellcolor[HTML]{ECF4FF}69.2 & \cellcolor[HTML]{ECF4FF}60.7          & \cellcolor[HTML]{ECF4FF}\textbf{66.8} & \cellcolor[HTML]{ECF4FF}57.4          & \cellcolor[HTML]{ECF4FF}\textbf{34.5} & \cellcolor[HTML]{ECF4FF}36.8          & \cellcolor[HTML]{ECF4FF}\textbf{80.1} & \cellcolor[HTML]{ECF4FF}\textbf{37.8} \\
\multirow{-6}{*}{\begin{tabular}[c]{@{}c@{}}4.5B\\ (35\%\\ Pruned)\end{tabular}} & \cellcolor[HTML]{ECF4FF}Ours: PPL   & \cellcolor[HTML]{ECF4FF}\textbf{23.1} & \cellcolor[HTML]{ECF4FF}\textbf{38.8} & \multicolumn{1}{c|}{\cellcolor[HTML]{ECF4FF}55.2}          & \cellcolor[HTML]{ECF4FF}64.3          & \cellcolor[HTML]{ECF4FF}71.4 & \cellcolor[HTML]{ECF4FF}59.4          & \cellcolor[HTML]{ECF4FF}59.3          & \cellcolor[HTML]{ECF4FF}\textbf{62.2} & \cellcolor[HTML]{ECF4FF}32.8          & \cellcolor[HTML]{ECF4FF}37.0          & \cellcolor[HTML]{ECF4FF}\textbf{80.1} & \cellcolor[HTML]{ECF4FF}\textbf{37.8} \\ 

\specialrule{.2em}{.1em}{.1em}

\specialrule{.2em}{.1em}{.1em} 

\multicolumn{2}{c|}{}                                                                                                  & \multicolumn{2}{c|}{PPL↓}                                                     & \multicolumn{8}{c|}{Commonsense Reasoning Accuracy↑ (\%)}                                                                                                                                                                                                                                                                        & \multicolumn{2}{c}{Thr↑ (tokens/s)\textsuperscript{$\ddagger$}}                                    \\
\multicolumn{2}{c|}{\multirow{-2}{*}{\#Param \& Method}}                                                                           & Wiki2                                 & PTB                                   & \multicolumn{1}{c|}{Average}                               & BoolQ                                 & PIQA                         & HellaSwag                             & WinoGrande                            & ARC-e                                 & ARC-c                                 & OBQA                         & H100                                  & RTX3090                               \\ \hline
\multicolumn{2}{c|}{Vicuna-7B: 6.7B}                                                                                   & 17.1                                  & 63.2                                  & \multicolumn{1}{c|}{65.9}                                  & 78.1                                  & 77.3                         & 73.9                                  & 69.5                                  & 74.3                                  & 44.3                                  & 43.8                         & 53.7                                  & 25.0                                  \\ \hline
                                                                                 & Wanda-sp                            & 24.4                                  & 104.0                                 & \multicolumn{1}{c|}{58.5}                                  & 63.9                                  & 72.0                         & 67.4                                  & 65.2                                  & 64.8                                  & 38.3                                  & 37.8                         & 41.7                                  & 16.7                                  \\
                                                                                 & FLAP                                & 22.0                                  & 74.9                                  & \multicolumn{1}{c|}{61.4}                                  & 73.1                                  & 74.8                         & 67.9                                  & 65.8                                  & 67.5                                  & \textbf{40.2}                         & \textbf{40.6}                & 40.5                                  & 16.5                                  \\
                                                                                 & LLM-Pruner                          & 19.6                                  & 76.4                                  & \multicolumn{1}{c|}{60.1}                                  & 65.4                                  & \textbf{76.2}                & 68.9                                  & 64.4                                  & 68.9                                  & 37.4                                  & 39.4                         & 43.2                                  & 21.4                                  \\
                                                                                 & SLEB                                & 25.1                                  & 77.0                                  & \multicolumn{1}{c|}{55.6}                                  & 63.2                                  & 72.1                         & 61.2                                  & 59.4                                  & 64.3                                  & 34.1                                  & 35.2                         & \textbf{66.0}                         & \textbf{28.4}                         \\
                                                                                 & \cellcolor[HTML]{ECF4FF}Ours: Grad+ & \cellcolor[HTML]{ECF4FF}21.0          & \cellcolor[HTML]{ECF4FF}72.3          & \multicolumn{1}{c|}{\cellcolor[HTML]{ECF4FF}\textbf{62.5}} & \cellcolor[HTML]{ECF4FF}\textbf{78.7} & \cellcolor[HTML]{ECF4FF}74.8 & \cellcolor[HTML]{ECF4FF}\textbf{69.4} & \cellcolor[HTML]{ECF4FF}\textbf{68.5} & \cellcolor[HTML]{ECF4FF}68.2          & \cellcolor[HTML]{ECF4FF}38.7          & \cellcolor[HTML]{ECF4FF}39.6 & \cellcolor[HTML]{ECF4FF}\textbf{66.0} & \cellcolor[HTML]{ECF4FF}\textbf{28.4} \\
\multirow{-6}{*}{\begin{tabular}[c]{@{}c@{}}5.5B\\ (20\%\\ Pruned)\end{tabular}} & \cellcolor[HTML]{ECF4FF}Ours: PPL   & \cellcolor[HTML]{ECF4FF}\textbf{18.8} & \cellcolor[HTML]{ECF4FF}\textbf{67.9} & \multicolumn{1}{c|}{\cellcolor[HTML]{ECF4FF}60.7}          & \cellcolor[HTML]{ECF4FF}71.7          & \cellcolor[HTML]{ECF4FF}74.4 & \cellcolor[HTML]{ECF4FF}67.6          & \cellcolor[HTML]{ECF4FF}63.6          & \cellcolor[HTML]{ECF4FF}\textbf{69.3} & \cellcolor[HTML]{ECF4FF}38.9          & \cellcolor[HTML]{ECF4FF}39.4 & \cellcolor[HTML]{ECF4FF}\textbf{66.0} & \cellcolor[HTML]{ECF4FF}\textbf{28.4} \\ \hline
                                                                                 & Wanda-sp                            & 36.5                                  & 177.6                                 & \multicolumn{1}{c|}{50.9}                                  & 49.0                                  & 67.1                         & 57.2                                  & 59.2                                  & 57.6                                  & 33.7                                  & 32.4                         & 41.7                                  & 16.0                                  \\
                                                                                 & FLAP                                & 27.9                                  & 88.3                                  & \multicolumn{1}{c|}{57.1}                                  & 72.0                                  & 71.5                         & 62.0                                  & 61.2                                  & 61.2                                  & 35.4                                  & 36.6                         & 40.2                                  & 16.5                                  \\
                                                                                 & LLM-Pruner                          & \textbf{22.7}                         & 87.9                                  & \multicolumn{1}{c|}{57.1}                                  & 60.8                                  & \textbf{74.3}                & \textbf{65.9}                         & 60.9                                  & \textbf{64.4}                         & 34.6                                  & \textbf{38.8}                & 44.0                                  & 22.9                                  \\
                                                                                 & SLEB                                & 34.0                                  & 98.0                                  & \multicolumn{1}{c|}{49.9}                                  & 47.9                                  & 68.7                         & 54.6                                  & 56.1                                  & 58.4                                  & 31.3                                  & 32.4                         & \textbf{73.9}                         & \textbf{34.9}                         \\
                                                                                 & \cellcolor[HTML]{ECF4FF}Ours: Grad+ & \cellcolor[HTML]{ECF4FF}29.8          & \cellcolor[HTML]{ECF4FF}92.0          & \multicolumn{1}{c|}{\cellcolor[HTML]{ECF4FF}\textbf{60.2}} & \cellcolor[HTML]{ECF4FF}\textbf{78.8} & \cellcolor[HTML]{ECF4FF}71.8 & \cellcolor[HTML]{ECF4FF}64.4          & \cellcolor[HTML]{ECF4FF}\textbf{67.7} & \cellcolor[HTML]{ECF4FF}64.3          & \cellcolor[HTML]{ECF4FF}\textbf{36.4} & \cellcolor[HTML]{ECF4FF}37.6 & \cellcolor[HTML]{ECF4FF}\textbf{73.9} & \cellcolor[HTML]{ECF4FF}\textbf{34.9} \\
\multirow{-6}{*}{\begin{tabular}[c]{@{}c@{}}4.9B\\ (27\%\\ Pruned)\end{tabular}} & \cellcolor[HTML]{ECF4FF}Ours: PPL   & \cellcolor[HTML]{ECF4FF}23.0          & \cellcolor[HTML]{ECF4FF}\textbf{78.2} & \multicolumn{1}{c|}{\cellcolor[HTML]{ECF4FF}56.1}          & \cellcolor[HTML]{ECF4FF}66.4          & \cellcolor[HTML]{ECF4FF}72.9 & \cellcolor[HTML]{ECF4FF}60.6          & \cellcolor[HTML]{ECF4FF}59.2          & \cellcolor[HTML]{ECF4FF}63.1          & \cellcolor[HTML]{ECF4FF}33.8          & \cellcolor[HTML]{ECF4FF}37.0 & \cellcolor[HTML]{ECF4FF}\textbf{73.9} & \cellcolor[HTML]{ECF4FF}\textbf{34.9} \\ \hline
                                                                                 & Wanda-sp                            & 73.2                                  & 386.5                                 & \multicolumn{1}{c|}{39.4}                                  & 43.1                                  & 58.4                         & 36.3                                  & 53.3                                  & 34.5                                  & 23.7                                  & 26.4                         & 41.6                                  & 16.1                                  \\
                                                                                 & FLAP                                & 34.6                                  & 104.8                                 & \multicolumn{1}{c|}{53.7}                                  & 65.1                                  & 68.1                         & 57.0                                  & 63.1                                  & 56.9                                  & 32.0                                  & 34.0                         & 40.5                                  & 15.8                                  \\
                                                                                 & LLM-Pruner                          & 27.6                                  & 102.0                                 & \multicolumn{1}{c|}{53.5}                                  & 52.0                                  & \textbf{72.4}                & \textbf{61.6}                         & 59.9                                  & 58.0                                  & \textbf{33.3}                         & \textbf{37.0}                & 44.4                                  & 21.1                                  \\
                                                                                 & SLEB                                & 43.5                                  & 117.3                                 & \multicolumn{1}{c|}{45.4}                                  & 41.3                                  & 65.9                         & 47.3                                  & 51.5                                  & 51.6                                  & 28.0                                  & 32.2                         & \textbf{80.1}                         & \textbf{37.8}                         \\
                                                                                 & \cellcolor[HTML]{ECF4FF}Ours: Grad+ & \cellcolor[HTML]{ECF4FF}35.0          & \cellcolor[HTML]{ECF4FF}110.3         & \multicolumn{1}{c|}{\cellcolor[HTML]{ECF4FF}\textbf{55.0}} & \cellcolor[HTML]{ECF4FF}64.0          & \cellcolor[HTML]{ECF4FF}69.6 & \cellcolor[HTML]{ECF4FF}59.3          & \cellcolor[HTML]{ECF4FF}\textbf{66.5} & \cellcolor[HTML]{ECF4FF}57.5          & \cellcolor[HTML]{ECF4FF}\textbf{33.3} & \cellcolor[HTML]{ECF4FF}35.2 & \cellcolor[HTML]{ECF4FF}\textbf{80.1} & \cellcolor[HTML]{ECF4FF}\textbf{37.8} \\
\multirow{-6}{*}{\begin{tabular}[c]{@{}c@{}}4.5B\\ (35\%\\ Pruned)\end{tabular}} & \cellcolor[HTML]{ECF4FF}Ours: PPL   & \cellcolor[HTML]{ECF4FF}\textbf{26.6} & \cellcolor[HTML]{ECF4FF}\textbf{89.4} & \multicolumn{1}{c|}{\cellcolor[HTML]{ECF4FF}53.3}          & \cellcolor[HTML]{ECF4FF}\textbf{65.2} & \cellcolor[HTML]{ECF4FF}70.4 & \cellcolor[HTML]{ECF4FF}56.5          & \cellcolor[HTML]{ECF4FF}56.6          & \cellcolor[HTML]{ECF4FF}\textbf{59.8} & \cellcolor[HTML]{ECF4FF}31.5          & \cellcolor[HTML]{ECF4FF}33.4 & \cellcolor[HTML]{ECF4FF}\textbf{80.1} & \cellcolor[HTML]{ECF4FF}\textbf{37.8} \\ 

\specialrule{.2em}{.1em}{.1em}

\specialrule{.2em}{.1em}{.1em} 

\multicolumn{2}{c|}{}                                                                                                   & \multicolumn{2}{c|}{PPL↓}                                                     & \multicolumn{8}{c|}{Commonsense Reasoning Accuracy↑ (\%)}                                                                                                                                                                                                                                                                        & \multicolumn{2}{c}{Thr↑ (tokens/s)\textsuperscript{$\ddagger$}}                                           \\
\multicolumn{2}{c|}{\multirow{-2}{*}{\#Param \& Method}}                                                                            & Wiki2                                 & PTB                                   & \multicolumn{1}{c|}{Average}                               & BoolQ                                 & PIQA                         & HellaSwag                             & WinoGrande                            & ARC-e                                 & ARC-c                                 & OBQA                         & H100                                  & RTX3090                               \\ \hline
\multicolumn{2}{c|}{Vicuna-13B: 13.0B}                                                                                  & 14.7                                  & 51.6                                  & \multicolumn{1}{c|}{68.3}                                  & 82.8                                  & 78.3                         & 77.0                                  & 71.2                                  & 75.4                                  & 47.7                                  & 45.4                         & 45.5                                  & OOM                                   \\ \hline
                                                                                  & Wanda-sp                            & 19.0                                  & 71.8                                  & \multicolumn{1}{c|}{63.6}                                  & 78.6                                  & 75.6                         & 73.5                                  & 68.4                                  & 68.5                                  & 42.2                                  & 38.4                         & 34.1                                  & 12.9                                  \\
                                                                                  & FLAP                                & 18.8                                  & 65.3                                  & \multicolumn{1}{c|}{63.3}                                  & 77.2                                  & 75.1                         & 72.0                                  & 70.2                                  & 69.4                                  & 40.3                                  & 38.8                         & 32.6                                  & 12.6                                  \\
                                                                                  & LLM-Pruner                          & \textbf{16.0}                         & 57.0                                  & \multicolumn{1}{c|}{65.3}                                  & 75.5                                  & \textbf{78.6}                & 75.0                                  & 69.8                                  & 70.6                                  & 43.6                                  & \textbf{44.4}                & 34.0                                  & 17.3                                  \\
                                                                                  & SLEB                                & 20.5                                  & 68.7                                  & \multicolumn{1}{c|}{60.4}                                  & 71.3                                  & 73.4                         & 68.3                                  & 63.9                                  & 66.8                                  & 38.7                                  & 40.2                         & \textbf{55.7}                         & \textbf{23.9}                         \\
                                                                                  & \cellcolor[HTML]{ECF4FF}Ours: Grad+ & \cellcolor[HTML]{ECF4FF}18.1          & \cellcolor[HTML]{ECF4FF}61.6          & \multicolumn{1}{c|}{\cellcolor[HTML]{ECF4FF}\textbf{66.7}} & \cellcolor[HTML]{ECF4FF}\textbf{83.0} & \cellcolor[HTML]{ECF4FF}76.8 & \cellcolor[HTML]{ECF4FF}\textbf{75.1} & \cellcolor[HTML]{ECF4FF}\textbf{72.8} & \cellcolor[HTML]{ECF4FF}\textbf{72.5} & \cellcolor[HTML]{ECF4FF}\textbf{44.5} & \cellcolor[HTML]{ECF4FF}42.4 & \cellcolor[HTML]{ECF4FF}\textbf{55.7} & \cellcolor[HTML]{ECF4FF}\textbf{23.9} \\
\multirow{-6}{*}{\begin{tabular}[c]{@{}c@{}}10.5B\\ (21\%\\ Pruned)\end{tabular}} & \cellcolor[HTML]{ECF4FF}Ours: PPL   & \cellcolor[HTML]{ECF4FF}16.1          & \cellcolor[HTML]{ECF4FF}\textbf{56.5} & \multicolumn{1}{c|}{\cellcolor[HTML]{ECF4FF}64.9}          & \cellcolor[HTML]{ECF4FF}75.0          & \cellcolor[HTML]{ECF4FF}77.1 & \cellcolor[HTML]{ECF4FF}73.7          & \cellcolor[HTML]{ECF4FF}68.9          & \cellcolor[HTML]{ECF4FF}71.5          & \cellcolor[HTML]{ECF4FF}43.8          & \cellcolor[HTML]{ECF4FF}44.2 & \cellcolor[HTML]{ECF4FF}\textbf{55.7} & \cellcolor[HTML]{ECF4FF}\textbf{23.9} \\ \hline
                                                                                  & Wanda-sp                            & 23.4                                  & 84.9                                  & \multicolumn{1}{c|}{60.0}                                  & 71.5                                  & 74.2                         & 68.7                                  & 65.1                                  & 64.3                                  & 36.8                                  & 39.4                         & 33.7                                  & 13.5                                  \\
                                                                                  & FLAP                                & 22.8                                  & 78.8                                  & \multicolumn{1}{c|}{61.6}                                  & 75.9                                  & 73.7                         & 67.9                                  & 66.4                                  & 67.3                                  & 38.0                                  & 42.0                         & 33.0                                  & 12.1                                  \\
                                                                                  & LLM-Pruner                          & 19.0                                  & 66.4                                  & \multicolumn{1}{c|}{62.7}                                  & 68.3                                  & \textbf{77.1}                & 72.0                                  & 69.7                                  & 68.6                                  & 40.0                                  & \textbf{43.4}                & 35.8                                  & 15.0                                  \\
                                                                                  & SLEB                                & 26.2                                  & 85.0                                  & \multicolumn{1}{c|}{56.0}                                  & 61.3                                  & 71.4                         & 64.1                                  & 59.6                                  & 60.0                                  & 37.0                                  & 38.4                         & \textbf{62.0}                         & \textbf{24.2}                         \\
                                                                                  & \cellcolor[HTML]{ECF4FF}Ours: Grad+ & \cellcolor[HTML]{ECF4FF}22.0          & \cellcolor[HTML]{ECF4FF}70.3          & \multicolumn{1}{c|}{\cellcolor[HTML]{ECF4FF}\textbf{65.1}} & \cellcolor[HTML]{ECF4FF}\textbf{82.6} & \cellcolor[HTML]{ECF4FF}75.1 & \cellcolor[HTML]{ECF4FF}\textbf{73.3} & \cellcolor[HTML]{ECF4FF}\textbf{70.9} & \cellcolor[HTML]{ECF4FF}69.9          & \cellcolor[HTML]{ECF4FF}\textbf{43.8} & \cellcolor[HTML]{ECF4FF}40.2 & \cellcolor[HTML]{ECF4FF}\textbf{62.0} & \cellcolor[HTML]{ECF4FF}\textbf{24.2} \\
\multirow{-6}{*}{\begin{tabular}[c]{@{}c@{}}9.5B\\ (29\%\\ Pruned)\end{tabular}}  & \cellcolor[HTML]{ECF4FF}Ours: PPL   & \cellcolor[HTML]{ECF4FF}\textbf{18.1} & \cellcolor[HTML]{ECF4FF}\textbf{62.2} & \multicolumn{1}{c|}{\cellcolor[HTML]{ECF4FF}62.0}          & \cellcolor[HTML]{ECF4FF}67.5          & \cellcolor[HTML]{ECF4FF}75.6 & \cellcolor[HTML]{ECF4FF}70.6          & \cellcolor[HTML]{ECF4FF}65.5          & \cellcolor[HTML]{ECF4FF}\textbf{70.9} & \cellcolor[HTML]{ECF4FF}43.3          & \cellcolor[HTML]{ECF4FF}40.2 & \cellcolor[HTML]{ECF4FF}\textbf{62.0} & \cellcolor[HTML]{ECF4FF}\textbf{24.2} \\ \hline
                                                                                  & Wanda-sp                            & 36.6                                  & 123.5                                 & \multicolumn{1}{c|}{52.7}                                  & 59.6                                  & 67.5                         & 59.5                                  & 59.7                                  & 55.2                                  & 33.5                                  & 33.8                         & 33.8                                  & 12.6                                  \\
                                                                                  & FLAP                                & 28.7                                  & 96.2                                  & \multicolumn{1}{c|}{58.3}                                  & 72.5                                  & 70.0                         & 62.5                                  & 65.4                                  & 63.8                                  & 36.3                                  & 37.8                         & 32.9                                  & 13.2                                  \\
                                                                                  & LLM-Pruner                          & 22.2                                  & 74.0                                  & \multicolumn{1}{c|}{59.7}                                  & 67.1                                  & \textbf{75.6}                & 67.7                                  & 63.2                                  & \textbf{65.5}                         & 38.8                                  & \textbf{39.8}                & 35.6                                  & 18.0                                  \\
                                                                                  & SLEB                                & 41.6                                  & 116.5                                 & \multicolumn{1}{c|}{49.4}                                  & 47.8                                  & 67.8                         & 54.5                                  & 56.1                                  & 53.8                                  & 32.2                                  & 33.6                         & \textbf{69.7}                         & \textbf{31.7}                         \\
                                                                                  & \cellcolor[HTML]{ECF4FF}Ours: Grad+ & \cellcolor[HTML]{ECF4FF}34.2          & \cellcolor[HTML]{ECF4FF}90.4          & \multicolumn{1}{c|}{\cellcolor[HTML]{ECF4FF}\textbf{61.4}} & \cellcolor[HTML]{ECF4FF}\textbf{78.5} & \cellcolor[HTML]{ECF4FF}71.3 & \cellcolor[HTML]{ECF4FF}\textbf{69.2} & \cellcolor[HTML]{ECF4FF}\textbf{69.9} & \cellcolor[HTML]{ECF4FF}64.2          & \cellcolor[HTML]{ECF4FF}\textbf{40.5} & \cellcolor[HTML]{ECF4FF}36.6 & \cellcolor[HTML]{ECF4FF}\textbf{69.7} & \cellcolor[HTML]{ECF4FF}\textbf{31.7} \\
\multirow{-6}{*}{\begin{tabular}[c]{@{}c@{}}8.3B\\ (37\%\\ Pruned)\end{tabular}}  & \cellcolor[HTML]{ECF4FF}Ours: PPL   & \cellcolor[HTML]{ECF4FF}\textbf{22.1} & \cellcolor[HTML]{ECF4FF}\textbf{73.6} & \multicolumn{1}{c|}{\cellcolor[HTML]{ECF4FF}59.1}          & \cellcolor[HTML]{ECF4FF}69.4          & \cellcolor[HTML]{ECF4FF}73.8 & \cellcolor[HTML]{ECF4FF}64.4          & \cellcolor[HTML]{ECF4FF}62.5          & \cellcolor[HTML]{ECF4FF}65.1          & \cellcolor[HTML]{ECF4FF}39.2          & \cellcolor[HTML]{ECF4FF}39.0 & \cellcolor[HTML]{ECF4FF}\textbf{69.7} & \cellcolor[HTML]{ECF4FF}\textbf{31.7} \\ 

\specialrule{.2em}{.1em}{.1em}

\end{tabular}
\begin{tablenotes}[para,flushleft]
\footnotesize
\textsuperscript{$\ddagger$}Throughput measured with 12 input tokens, 128 output tokens, and a batch size of 1 on a single GPU.
\end{tablenotes}
\end{threeparttable}
\end{adjustbox}
\vspace{-0.06in}
\caption{Results of pruned LLaMA-7B (top), Vicuna-7B-v1.3 (middle), and Vicuna-13B-v1.3 (bottom). The width pruning of Wanda-sp~\cite{wanda,flap}, FLAP~\cite{flap}, and LLM-Pruner~\cite{llmpruner} often degrades inference efficiency due to the GPU-unfriendly weight sizes~\cite{tensor_core_guide}. In contrast, our depth pruning delivers actual speedups while performing comparably with light LoRA retraining.} \label{supple_lora_results}
\end{table*}

\subsection{Generation Examples}\label{sec:supple_gen_ex}
Tables~\ref{supple_genEx_7b}~and~\ref{supple_genEx_13b} present generation examples where some input prompts were sourced from~\citet{llmpruner}. In terms of linguistic flow and topical consistency, the pruned models yield sentences on par with those from the original model. However, as also noted in~\citet{llmpruner}, the output quality deteriorates when responding to factual questions or producing longer content. To overcome this challenge, it is essential to apply a more powerful retraining method on a large-scale corpus.
 
\begin{table*}[h!]
\centering
\vspace{-0.05in}
\begin{adjustbox}{max width=0.97\columnwidth}
\begin{threeparttable}
\begin{tabular}{p{1.8cm}|p{14.5cm}}

\specialrule{.2em}{.1em}{.1em} 
Model       & Generated Text 
\\ \hline

\small{LLaMA-7B}    & 
\footnotesize \uline{What’s great about the holiday season,} it’s a time when people focus less on work and more on families and friends. The holidays are a time when people can really focus on giving and helping others, which is something that the holidays are supposed to be about. In my blog post this week, I’ll be taking a look at a variety of causes that ...
\\ \hline

\small{Wanda-sp}    & 
\footnotesize \uline{What’s great about the holiday season,} though, is its warm and inviting people, and what a chance to meet and share a cuppa with others who understand and have shared similar experiences, a feeling of empathy if you will, and a great opportunity to help others on a difficult journey. So I can say, all is well. But I'm not giving up the fight ...
\\ \hline

 \small{FLAP}        & 
\footnotesize \uline{What’s great about the holiday season,} and what makes it such an awesome time of the year, is that it’s the time of the year when people are so generous, and so goodhearted. When you get together with friends and family, and you go through the various things that you’re able to do, when you’re able to give people gifts, when people can see ...
\\ \hline

 \small{LLM-Pruner}  & 
\footnotesize \uline{What’s great about the holiday season,} is you can enjoy the company of friends and family. This is the perfect opportunity to gather together, and share all the special memories that you have together. These moments are the ones that make the holiday special. No matter what happens, or what you end up doing, it’s always good to get ...
\\ \hline

 \small{Ours: Taylor+} & 
\footnotesize \uline{What’s great about the holiday season,} is that it’s a time of year when people can actually take a step back from the normal day-to-day routines and take some time out to relax and enjoy some well-deserved rest and rejuvenation. This might include indulging in a more comfortable or luxury holiday experience and this article aims to help  ...
 \\ \hline
 
 \small{Ours: PPL}   & 
\footnotesize \uline{What’s great about the holiday season,} in my mind, is that it brings together friends, family, and food. The holiday season is a great time to show some love and have some fun with the people that matter most. Whether you are celebrating Christmas, Hanukah, or New Years, the festivities will be full of joy and memories that will last ...
\\ 

\specialrule{.2em}{.1em}{.1em}

\specialrule{.2em}{.1em}{.1em}

\small{LLaMA-7B}    & 
\footnotesize \uline{AI can create a logo in seconds}. Why are these new technologies threatening our jobs? AI can do most menial tasks in just a few seconds or minutes. These tasks can be very tedious for human beings to do. For instance, AI can identify patterns in large amounts of data so that complex statistical analysis can be carried out faster than a ...
\\ \hline

\small{Wanda-sp}    & 
\footnotesize \uline{AI can create a logo in seconds} rather than years. To this point, this is only the beginning of what AI will do to change our human lives. As time goes on, it’s quite likely that we’ll see even greater use of robots. Even if this all sounds too dramatic, this may still happen. And AI is the best thing that can make it to happen. in 2019. All you ...
\\ \hline

 \small{FLAP}        & 
\footnotesize \uline{AI can create a logo in seconds} and it doesn’t cost you a penny. The new era of technology has opened up many exciting opportunities. Technology, like AI is being utilized for different purposes in different ways, and now, technology is becoming a solution to problems that would have been tedious or expensive only a few years ago. ...
\\ \hline
 \small{LLM-Pruner}  & 
\footnotesize \uline{AI can create a logo in seconds}, design a website from scratch and even write blog posts without any human intervention. This revolution is changing the way humans communicate and express ideas. In our view, this technology is a great asset but we'll need to be smart in how we use it to prevent abuse or automation leading to ...
\\ \hline

 \small{Ours: Taylor+} & 
\footnotesize \uline{AI can create a logo in seconds} or design a new website. There is a lot of software that exists in these two categories. It has become easier and faster for people to use. However, there is still the requirement for the designer or creative director to ensure that the designs are relevant and meaningful to the brand. The Future: AI for creative design ...
\\ \hline
 \small{Ours: PPL}   & 
\footnotesize \uline{AI can create a logo in seconds}, and it can design entire websites, all without human intervention. However, there is still a lot of work to be done to perfect this technology, including the development of natural language processing, machine learning and computer vision, as well as ensuring that AI designers are creating work that can be ...
\\ 

\specialrule{.2em}{.1em}{.1em}

\specialrule{.2em}{.1em}{.1em}

\small{LLaMA-7B}    & 
\footnotesize \uline{Neural network pruning is defined as} the process of removing certain weights or neurons in a neural network to reduce the size of the neural network, and improve the training process. Neural networks pruning techniques have been applied successfully on classification, optimization, planning, and speech recognition tasks. A pruned ...
\\ \hline
\small{Wanda-sp}    & 
\footnotesize \uline{Neural network pruning is defined as} any method that removes the redundant information by applying the neural network network. It is the term used to identify the systematic variables when an individual tries to find an algorithm. In addition, it is the method of classification used to define any set of algorithms that have been ...
\\ \hline
 \small{FLAP}        & 
\footnotesize \uline{Neural network pruning is defined as} a selection of neurons or weights in a neural network so that the information of the network is reduced to a specified level, and a process of re-training a neural network. The latter one is the most popular method, especially the neural network pruning. In recent years, this field has been studied in ...
\\ \hline
 \small{LLM-Pruner}  & 
\footnotesize \uline{Neural network pruning is defined as} reducing the complexity and size of an over-parameterized neural network. The pruning reduces the FLOPs and improves performance for the particular task. Pruning techniques have gained popularity in recent years due to their wide range of applications. This paper aims to explore the pruning ...
\\ \hline

 \small{Ours: Taylor+} & 
\footnotesize \uline{Neural network pruning is defined as} reducing the weight connections and biases of a neural network, while retaining high performance. Pruning is an effective approach to reducing the number of parameters and improving the efficiency of neural networks. Recently, the concept of neural network shrinking has gained considerable ...
\\ \hline
 \small{Ours: PPL}   & 
\footnotesize \uline{Neural network pruning is defined as} the process of removing redundant neurons and connections between nodes in a neural network, while keeping the performance or accuracy to the maximum extent possible. This optimization is aimed at reducing the size and memory consumption of the network while maintaining its ability to perform task ...
\\ 

\specialrule{.2em}{.1em}{.1em}
\end{tabular}
\end{threeparttable}
\end{adjustbox}
\vspace{-0.05in}
\caption{Generation examples from the original LLaMA-7B and 20\%-compressed models.} \label{supple_genEx_7b}
\end{table*}

\begin{table*}[ht!]
\centering
\vspace{-0.05in}
\begin{adjustbox}{max width=0.97\columnwidth}
\begin{threeparttable}
\begin{tabular}{p{1.8cm}|p{14.5cm}}

\specialrule{.2em}{.1em}{.1em} 
Model       & Generated Text 
\\ \hline

 \small{Vicuna-13B}    & 
\footnotesize \uline{What’s great about the holiday season,} beyond the obvious time off and the chance to spend time with loved ones, is the opportunity to slow down and reflect on the year that’s passed. It’s a time to look back at all the accomplishments and growth you’ve experienced, and to set your sights on the year ahead. As a small business owner, the ...
\\ \hline

 \small{Wanda-sp}    & 
\footnotesize \uline{What’s great about the holiday season,} besides spending time with family and friends, is that it’s a time of giving. It’s a time to show appreciation for those in your life who make a difference and bring joy to your world. Whether it’s a gift of love, a gift of time, or a gift of money, it’s important to remember that giving is not just about ...
\\ \hline

\small{FLAP}        & 
\footnotesize \uline{What’s great about the holiday season,} especially the days leading up to Christmas, is the opportunity to take a pause and recharge. It’s a time to reflect on the year and all the experiences we’ve had, good or bad. It’s a time to reflect on the people we love and the impact they’ve had on our lives. It’s a time to reflect on what we want ...
\\ \hline

\small{LLM-Pruner}  & 
\footnotesize \uline{What’s great about the holiday season,} aside from the gifts and the holiday parties, is the opportunity to spend time with family and loved ones. One of the things that I love most about holiday seasons is the traditions that surround them, from decorating the tree to making special holiday meals, there’s a sense of nostalgia and warmth that ...
\\ \hline

 \small{Ours: Taylor+} & 
\footnotesize \uline{What’s great about the holiday season,} is that it allows us to reflect on the year gone by and reflect on the moments that made it special. And while some people might choose to celebrate with the traditional traditions, some choose to do something different. One way to spice up your holiday season is by taking a trip to a far-off place. Why not ...
\\ \hline 
 \small{Ours: PPL}   & 
\footnotesize \uline{What’s great about the holiday season,} in addition to the fun food, festivities, and spending time with loved ones, is that it creates the perfect opportunity to show people how much you appreciate them. From giving them a gift, to baking their favorite treat, or just taking a moment to acknowledge them in front of others – there’s something ...
\\ 
\specialrule{.2em}{.1em}{.1em}

\specialrule{.2em}{.1em}{.1em}

 \small{Vicuna-13B}    & 
\footnotesize \uline{AI can create a logo in seconds} with its complex algorithms and pre-set designs. But it can never truly understand the essence of a business, the emotions that go into creating a logo or the creativity and uniqueness that comes from human input. It's time to put the human touch back into branding, where a designer and client can collaborate to ...
\\ \hline

 \small{Wanda-sp}    & 
\footnotesize \uline{AI can create a logo in seconds} using a logo design tool. However, a talented graphic designer can create a unique and meaningful logo that stands out from the crowd, captures the essence of the business, and appeals to the target audience. A good logo can also make a difference in how customers perceive the business and can even increase ...
\\ \hline

\small{FLAP}        & 
\footnotesize \uline{AI can create a logo in seconds}. If you don't like it, you can change it yourself or pay for a designer to do it for you. However, it's important to note that a logo is more than just an image. It's a visual representation of a brand or organization, and it communicates certain values and characteristics to the viewer. Therefore, it's essential to ...
\\ \hline
\small{LLM-Pruner}  & 
\footnotesize \uline{AI can create a logo in seconds} with just one click, but design is a human art, something that is beyond AI capabilities. The logo design is an important part of a company's identity, and it's a decision that affects the company's visual branding and reputation for years to come. The logo is the visual representation of a company's values ...
\\ \hline

 \small{Ours: Taylor+} & 
\footnotesize \uline{AI can create a logo in seconds}, with or without a human's input. Some people might like this, because they can create a logo for a small business on a budget. Other people, like myself, think that the human touch is important in art and design. A human designer can create a unique and meaningful logo by incorporating personal ...
\\ \hline
 \small{Ours: PPL}   & 
\footnotesize \uline{AI can create a logo in seconds} by combining a variety of different styles and shapes and patterns that make up the logos. However, a logo that is truly a reflection of your brand needs to be designed by a human, a graphic designer. They will bring a creative vision that will make your logo unique, appealing and memorable. They can add ...
\\ 
\specialrule{.2em}{.1em}{.1em}

\specialrule{.2em}{.1em}{.1em}

 \small{Vicuna-13B}    & 
\footnotesize \uline{Neural network pruning is defined as} the process of removing redundant or unnecessary connections in a neural network to reduce its size and computational complexity while preserving its performance. The primary motivation behind pruning is to improve the efficiency of deep learning models, especially in resource-constrained ...
\\ \hline
 \small{Wanda-sp}    & 
\footnotesize \uline{Neural network pruning is defined as} a method for removing connections in order to reduce the model size, thus allowing for computationally efficient models while retaining accuracy. This technique is useful in reducing the number of neurons in a neural network, which allows for fewer computations during training and reduces the ...
\\ \hline
\small{FLAP}        & 
\footnotesize \uline{Neural network pruning is defined as} the process of reducing the size of a neural network by removing the least significant weights or neurons. The main motivation for pruning is to reduce the computational cost of neural networks while maintaining their accuracy. The key challenge of pruning is to identify the most effective ...
\\ \hline
\small{LLM-Pruner}  & 
\footnotesize \uline{Neural network pruning is defined as} a technique that aims to reduce the size of a neural network by removing redundant and irrelevant connections between the neurons in the network. This approach is based on the observation that a large portion of the connections within the network is redundant and does not contribute to the overall ...
\\ \hline

 \small{Ours: Taylor+} & 
\footnotesize \uline{Neural network pruning is defined as} the removal of redundant connections within a neural network to achieve a better model fit while retaining the network's general accuracy. The goal of pruning is to reduce the computational cost and memory footprint of the network. One commonly used pruning method is called weight magnitude ...
\\ \hline
 \small{Ours: PPL}   & 
\footnotesize \uline{Neural network pruning is defined as} the task of removing unnecessary or redundant connections in a neural network while retaining its accuracy and performance. This is often done to reduce the memory usage and computational complexity of a neural network, which can be critical when running on devices with limited resources. In ...
\\ 

\specialrule{.2em}{.1em}{.1em}
\end{tabular}
\end{threeparttable}
\end{adjustbox}
\vspace{-0.05in}
\caption{Generation examples from the original Vicuna-13B-v1.3 and 21\%-compressed models.} \label{supple_genEx_13b}
\end{table*}

\clearpage
\section{Further Results of Aggressive Pruning and CPT Retraining}\label{sec:supple_aggressive_pruning}

\begin{figure}[h]
  \centering
    \includegraphics[width=\linewidth]{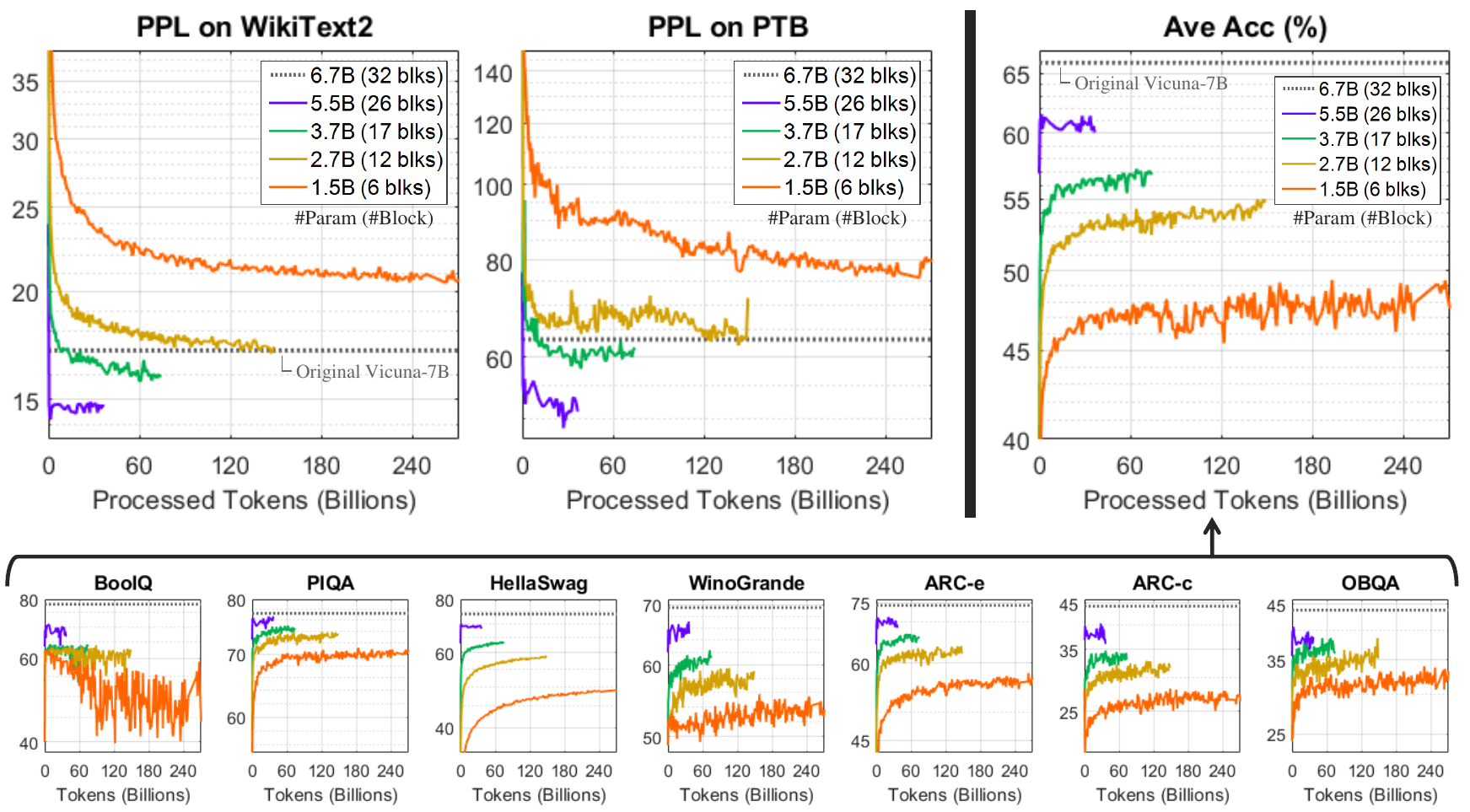}
        \vspace{-0.2in}
  \caption{Zero-shot performance over the course of training for models from Vicuna-7B-v1.3 at different pruning ratios. For each model size, the CPT duration was limited to a two-week period, but additional training could further improve the quality.} \label{fig_appendix_cpt_learn_curve}
\end{figure}

Figure~\ref{fig_appendix_cpt_learn_curve} illustrates the learning curve for models pruned from Vicuna-7B. For each model size, retraining is performed within a two-week span using eight NVIDIA H100 GPUs. The CPT of the 5.5B-parameter model shows early convergence, completing in just 6 days while processing 37B tokens, potentially owing to significant knowledge retained within the network. In contrast, the CPTs for the 3.7B, 2.7B, and 1.5B models take 8, 12, and 11 days, respectively, to process 74B, 150B, and 271B tokens. This process allows the models to restore lost capabilities while achieving improved inference efficiency. Longer training periods could further enhance the recovery of model quality.

\clearpage
\section{Compatibility with PTQ}\label{sec:appendix_gptq}

Our pruning approach can be combined with quantization to further decrease memory usage. To validate this aspect, we apply 4-bit GPTQ~\cite{frantar2023optq} to our pruned models, using 128 randomly sampled sequences with 2048 tokens from the C4 dataset~\cite{T5C4dataset2019} as calibration data for PTQ. The results demonstrate that quantization does not cause a noticeable degradation in zero-shot model performance while leading to additional computational reductions.

\subsection{Zero-shot Performance after Applying Quantization}
\begin{table*}[ht!]
\centering
\begin{adjustbox}{max width=\linewidth}
\begin{threeparttable}
\begin{tabular}{ccc|cc|cccccccc}
\specialrule{.2em}{.1em}{.1em}
\multicolumn{3}{c|}{Model} &
  \multicolumn{2}{c|}{PPL ↓} &
  \multicolumn{8}{c}{Commonsense Reasoning Accuracy↑ (\%)} \\ \hline
\#Param &
  Retraining &
  Quantization &
  Wiki2 &
  PTB &
  \multicolumn{1}{c|}{Average} &
  BoolQ &
  PIQA &
  HellaSwag &
  WinoGrande &
  ARC-e &
  ARC-c &
  OBQA \\ 
  \hline
\multirow{2}{*}{\begin{tabular}[c]{@{}c@{}}6.7B\\ (Original)\end{tabular}} &
  \multirow{2}{*}{-} &
  \xmark &
  17.1 &
  63.2 &
  \multicolumn{1}{c|}{65.9} &
  78.1 &
  77.3 &
  73.9 &
  69.5 &
  74.3 &
  44.3 &
  43.8 \\
 &
   &
  \cmark &
  17.3 &
  64.8 &
  \multicolumn{1}{c|}{63.6} &
  72.5 &
  76.4 &
  72.4 &
  67.6 &
  72.8 &
  42.7 &
  40.4 \\ 
  \hline \hline
\multirow{6}{*}{\begin{tabular}[c]{@{}c@{}}5.5B\\ (20\%\\ Pruned)\end{tabular}} &
  \multirow{2}{*}{LoRA} &
  \xmark &
  18.8 &
  67.9 &
  \multicolumn{1}{c|}{60.7} &
  71.7 &
  74.4 &
  67.6 &
  63.6 &
  69.3 &
  38.9 &
  39.4 \\
 &
   &
  \cmark &
  19.7 &
  70.7 &
  \multicolumn{1}{c|}{60.1} &
  70.2 &
  74.6 &
  66.9 &
  64.4 &
  67.6 &
  38.6 &
  38.4 \\ \cline{2-13} 
 &
  \multirow{2}{*}{CPT} &
  \xmark &
  14.3 &
  56.2 &
  \multicolumn{1}{c|}{61.5} &
  70.5 &
  75.7 &
  69.9 &
  65.7 &
  70.4 &
  39.2 &
  39.2 \\
 &
   &
  \cmark &
  15.1 &
  59.3 &
  \multicolumn{1}{c|}{60.6} &
  69.7 &
  75.9 &
  68.9 &
  63.9 &
  68.5 &
  38.5 &
  38.6 \\ \cline{2-13} 
 &
  \multirow{2}{*}{\begin{tabular}[c]{@{}c@{}}CPT$\Rightarrow$LoRA\end{tabular}} &
  \xmark &
  14.8 &
  60.2 &
  \multicolumn{1}{c|}{63.1} &
  72.5 &
  77.5 &
  71.1 &
  66.0 &
  72.1 &
  41.1 &
  41.0 \\
 &
   &
  \cmark &
  15.5 &
  64.1 &
  \multicolumn{1}{c|}{61.7} &
  71.1 &
  76.4 &
  70.3 &
  64.2 &
  71.5 &
  40.9 &
  37.6 \\  \hline \hline
\multirow{6}{*}{\begin{tabular}[c]{@{}c@{}}3.7B\\ (45\%\\ Pruned)\end{tabular}} & 
  \multirow{2}{*}{LoRA} &
  \xmark &
  37.0 &
  113.2 &
  \multicolumn{1}{c|}{47.0} &
  54.3 &
  67.1 &
  45.3 &
  53.4 &
  52.2 &
  27.6 &
  28.8 \\
 &
   &
  \cmark &
  38.0 &
  117.6 &
  \multicolumn{1}{c|}{46.8} &
  55.3 &
  66.2 &
  45.1 &
  53.5 &
  50.5 &
  27.6 &
  29.2 \\ \cline{2-13} 
 &
  \multirow{2}{*}{CPT} &
  \xmark &
  16.0 &
  60.0 &
  \multicolumn{1}{c|}{57.1} &
  62.6 &
  74.5 &
  63.5 &
  62.4 &
  66.0 &
  34.4 &
  36.4 \\
 &
   &
  \cmark &
  16.6 &
  61.5 &
  \multicolumn{1}{c|}{57.1} &
  63.8 &
  74.5 &
  62.7 &
  61.0 &
  65.8 &
  34.2 &
  37.8 \\ \cline{2-13} 
 &
  \multirow{2}{*}{\begin{tabular}[c]{@{}c@{}}CPT$\Rightarrow$LoRA\end{tabular}} &
  \xmark &
  16.5 &
  60.5 &
  \multicolumn{1}{c|}{57.4} &
  62.0 &
  74.9 &
  64.8 &
  61.7 &
  65.2 &
  34.1 &
  39.0 \\
 &
   &
  \cmark &
  17.0 &
  61.8 &
  \multicolumn{1}{c|}{56.9} &
  61.0 &
  74.5 &
  64.1 &
  61.8 &
  64.7 &
  34.1 &
  38.4 \\  \hline \hline
\multirow{6}{*}{\begin{tabular}[c]{@{}c@{}}2.7B\\ (60\%\\ Pruned)\end{tabular}} &
  \multirow{2}{*}{LoRA} &
  \xmark &
  68.9 &
  196.4 &
  \multicolumn{1}{c|}{40.1} &
  41.3 &
  61.0 &
  33.9 &
  53.0 &
  40.4 &
  25.2 &
  26.0 \\
 &
   &
  \cmark &
  71.5 &
  205.9 &
  \multicolumn{1}{c|}{40.1} &
  42.7 &
  60.4 &
  33.7 &
  52.6 &
  40.7 &
  24.9 &
  25.8 \\ \cline{2-13} 
 &
  \multirow{2}{*}{CPT} &
  \xmark &
  17.1 &
  63.1 &
  \multicolumn{1}{c|}{55.0} &
  61.8 &
  73.5 &
  58.6 &
  58.2 &
  62.4 &
  31.8 &
  38.6 \\
 &
   &
  \cmark &
  17.7 &
  64.7 &
  \multicolumn{1}{c|}{54.6} &
  61.9 &
  73.1 &
  58.4 &
  58.8 &
  62.5 &
  31.8 &
  35.6 \\ \cline{2-13} 
 &
  \multirow{2}{*}{\begin{tabular}[c]{@{}c@{}}CPT$\Rightarrow$LoRA\end{tabular}} &
  \xmark &
  17.8 &
  65.1 &
  \multicolumn{1}{c|}{55.0} &
  61.4 &
  73.9 &
  59.7 &
  58.0 &
  61.3 &
  32.3 &
  38.0 \\
 &
   &
  \cmark &
  18.4 &
  66.1 &
  \multicolumn{1}{c|}{55.0} &
  61.9 &
  73.8 &
  59.0 &
  58.3 &
  62.1 &
  32.0 &
  38.0 \\ \hline \hline
\multirow{6}{*}{\begin{tabular}[c]{@{}c@{}}1.5B\\ (80\%\\ Pruned)\end{tabular}} &
  \multirow{2}{*}{LoRA} &
  \xmark &
  1002.2 &
  1874.9 &
  \multicolumn{1}{c|}{37.1} &
  51.6 &
  53.5 &
  26.4 &
  49.3 &
  27.8 &
  27.5 &
  24.0 \\
 &
   &
  \cmark &
  1014.3 &
  1932.4 &
  \multicolumn{1}{c|}{37.5} &
  53.7 &
  53.3 &
  26.5 &
  50.0 &
  28.2 &
  26.5 &
  24.6 \\ \cline{2-13} 
 &
  \multirow{2}{*}{CPT} &
  \xmark &
  20.5 &
  77.4 &
  \multicolumn{1}{c|}{49.2} &
  53.5 &
  70.7 &
  48.9 &
  54.5 &
  56.7 &
  27.0 &
  33.0 \\
 &
   &
  \cmark &
  21.4 &
  80.0 &
  \multicolumn{1}{c|}{48.5} &
  48.9 &
  70.1 &
  48.8 &
  54.1 &
  55.7 &
  26.8 &
  35.0 \\ \cline{2-13} 
 &
  \multirow{2}{*}{\begin{tabular}[c]{@{}c@{}}CPT$\Rightarrow$LoRA\end{tabular}} &
  \xmark &
  21.1 &
  79.0 &
  \multicolumn{1}{c|}{49.0} &
  52.5 &
  70.7 &
  49.6 &
  52.7 &
  55.6 &
  28.0 &
  34.0 \\
 &
   &
  \cmark &
  21.8 &
  82.0 &
  \multicolumn{1}{c|}{48.6} &
  51.7 &
  70.2 &
  49.8 &
  52.7 &
  55.0 &
  27.6 &
  33.4 \\ \specialrule{.2em}{.1em}{.1em}

\end{tabular}
\end{threeparttable}
\end{adjustbox}

\caption{Zero-shot results from applying PTQ to various pruned and retrained models derived from Vicuna-7B-v1.3.} \label{supple_gptq_ppl_acc}
\end{table*}

\subsection{Further GPU Memory Reduction from Quantization}
\begin{table*}[ht!]
\centering
\begin{adjustbox}{max width=0.88\linewidth}
\begin{threeparttable}
\begin{tabular}{cc|cccc|cccc}
\specialrule{.2em}{.1em}{.1em}
\multicolumn{2}{c|}{Model} & \multicolumn{4}{c|}{$L$128}       & \multicolumn{4}{c}{$L$512}               \\ \hline
\# Param   & Quantization  & $M$1    & $M$16   & $M$64    & $M$256   & $M$1    & $M$16    & $M$64    & $M$256         \\ \hline
\multirow{2}{*}{\begin{tabular}[c]{@{}c@{}}6.7B\\ (Original)\end{tabular}}      & \xmark & 12.8GB & 16.0GB & 25.8GB & 65.0GB & 13.3GB & 25.0GB & 61.8GB & OOM \\
           & \cmark          & 4.8GB & 7.8GB & 17.7GB & 56.9GB & 5.3GB & 16.9GB & 53.7GB & OOM \\ \hline
\multirow{2}{*}{\begin{tabular}[c]{@{}c@{}}5.5B\\ (20\% Pruned)\end{tabular}} & \xmark & 10.5GB & 13.1GB & 21.1GB & 52.9GB & 10.9GB & 20.4GB & 50.4GB & OOM \\
           & \cmark          & 4.1GB & 6.6GB & 14.7GB & 46.5GB & 4.5GB & 14.0GB & 43.9GB & OOM \\ \hline
\multirow{2}{*}{\begin{tabular}[c]{@{}c@{}}3.7B\\ (45\% Pruned)\end{tabular}} & \xmark & 7.1GB  & 8.7GB  & 14.0GB & 34.9GB & 7.4GB  & 13.5GB & 33.2GB & OOM \\
           & \cmark          & 3.1GB & 4.8GB & 10.1GB & 31.0GB & 3.4GB & 9.6GB  & 29.2GB & OOM \\ \hline
\multirow{2}{*}{\begin{tabular}[c]{@{}c@{}}2.7B\\ (60\% Pruned)\end{tabular}} & \xmark & 5.1GB  & 6.3GB  & 10.1GB & 24.9GB & 5.3GB  & 9.7GB  & 23.6GB & OOM \\
           & \cmark          & 2.6GB & 3.8GB & 7.5GB  & 22.3GB & 2.8GB & 7.2GB  & 21.0GB & 76.4GB       \\ \hline
\multirow{2}{*}{\begin{tabular}[c]{@{}c@{}}1.5B\\ (80\% Pruned)\end{tabular}} & \xmark & 2.8GB  & 3.4GB  & 5.4GB  & 12.9GB & 2.9GB  & 5.1GB  & 12.1GB & 39.9GB       \\
           & \cmark          & 1.9GB & 2.5GB & 4.5GB  & 12.0GB & 2.0GB & 4.2GB  & 11.2GB & 39.0GB \\ \specialrule{.2em}{.1em}{.1em}
\end{tabular}%
\end{threeparttable}
\end{adjustbox}

\caption{VRAM reduction by applying quantization after using our pruning method. The results of the pruned Vicuna-7B models and their 4-bit weight-quantized counterparts are reported under varying sequence lengths ($L$) and batch sizes ($M$).}\label{supple_gptq_vram}

\end{table*}

\clearpage
\section{Experimental Setup}\label{sec:supple_setup}
  
\subsection{Baseline Methods} \label{sec:supple_baseline}
We primarily compare the two pruning units, focusing on `network width \textit{vs.} depth,' and also include a very recent depth pruning method in our analysis. The baseline methods are described below, where we use their official code for implementation. To ensure a fair comparison, we employ the same calibration dataset across all methods. Table~\ref{supple_table:arch} shows the pruned architectures under similar numbers of parameters.

\begin{enumerate}[itemsep=0em]
\setlength{\leftskip}{-0.2cm}

\item[$\circ$] LLM-Pruner~\citep{llmpruner} employs a Taylor-based importance metric to remove attention heads from MHA and intermediate neurons from FFN. Local pruning is performed to select removable groups within the same module while maintaining uniform dimensions across the examined blocks. Adhering to their practice, the first and last few blocks remain unpruned. Their pruned models and ours are identically retrained with LoRA.

\item[$\circ$] FLAP~\citep{flap} uses a fluctuation-based importance metric to explore the recoverability of feature maps after removing weight columns. Global pruning is applied, leading to different widths over distinct modules (see Table~\ref{supple_table:arch} for mean and standard deviation values). Instead of retraining, extra bias terms are added into pruned feature maps for performance restoration.

\item[$\circ$] Wanda-sp is presented in~\citet{flap} as a variant of Wanda~\citep{wanda} adjusted for structured pruning. The original metric was based on the product of weight magnitudes and input activation norms, which can be interpreted as addressing a local reconstruction objective. Wanda-sp extends this in a structured way while using common dimensions among different modules.

\item[$\circ$] SLEB~\cite{song2024sleb} prunes Transformer blocks in LLMs and has been introduced concurrently with our study. It uses a logit-based method to find unnecessary blocks, similar to our PPL criterion, and updates the importance scores after each block is removed. Although SLEB pursues a retraining-free setup, we observed that it fails to sustain adequate performance as the pruning ratio increases.

\setlength{\leftskip}{0pt}
\end{enumerate}

\subsection{Implementation Details} \label{sec:supple_impl_details}
Our implementation employs the Transformers library~\cite{wolf-etal-2020-transformers}. An NVIDIA A100 (80GB VRAM) GPU is used for the pruning and LoRA retraining phases. For CPT retraining, eight NVIDIA H100 (80GB) GPUs are utilized, with each model size trained in under two weeks. 

\begin{enumerate}[itemsep=0em]
\setlength{\leftskip}{-0.2cm}

\item[$\circ$] At the pruning phase, we assess the significance of Transformer blocks using a small calibration set (containing 10 samples from BookCorpus~\citep{Zhu_2015_ICCV} with a sequence length of 128). For the PPL-based criterion, the calibration samples are fed into networks with a single block removed, and this step is iterated across all the blocks in the target model. For the Taylor+ method, we feed the calibration data into the original network to collect backward-gradient matrices. The pruning is completed efficiently within 1 to 2 hours for the 7B- and 13B-sized models. 
 
\item[$\circ$] At the LoRA retraining phase, we apply a LoRA adapter~\citep{lora} to every projection weight matrix by following \citet{llmpruner}. We employ a LoRA rank of 8, a learning rate of 0.0001, and a batch size of 64 over 2 epochs. The retraining costs are notably low, with the entire process being executed on a single GPU. For example, retraining a 20\%-pruned model from 7B parameters takes about 2 hours and utilizes 22GB GPU memory, while a 21\%-pruned model from 13B parameters requires approximately 3 hours and 35GB VRAM.

\item[$\circ$] At the CPT retraining phase, we utilize the AdamW optimizer with ($\beta_1$, $\beta_2$) values of (0.9, 0.95), under a weight decay of 0.1 and a learning rate of 0.0001. A global batch size of 512 is used, with a micro-batch size of 2 for 32 gradient accumulation steps over 8 GPUs. Gradient clipping with a max norm value of 1 is applied. The CPT for the 5.5B-parameter model takes only 6 days (covering 37B tokens) due to early convergence. On the other hand, the CPT for the 3.7B, 2.7B, and 1.5B models takes 8 days (74B tokens), 12 days (150B tokens), and 11 days (271B tokens), respectively. Due to constrained resources, we restricted our CPT procedure to not exceed two weeks for each model size; however, extending the training duration could further improve performance.
 
\item[$\circ$] At the inference stage, we maintain the default configuration of the Transformers library, without using xFormers-optimized attention or advanced options.

\setlength{\leftskip}{0pt}
\end{enumerate}

\begin{table}[h!]
\centering
\begin{adjustbox}{max width=0.65\columnwidth}
\begin{threeparttable}
\begin{tabular}{cc|c|c|cc}
\specialrule{.2em}{.1em}{.1em} 

\multicolumn{2}{c|}{Model}                                                          & \#Param & \#Block\textsuperscript{$\ddagger$} & \#Head\textsuperscript{$\ddagger$}    & FFN-D\textsuperscript{$\ddagger$}          \\ \hline
\multicolumn{2}{c|}{Original 7B}                                                    & 6.7B    & 32      & 32        & 11008          \\ \hline
\multirow{4}{*}{\begin{tabular}[c]{@{}c@{}}20\%\\ Pruned\textsuperscript{$\dagger$}\end{tabular}} & Wanda-sp   & 5.5B    & 32      & 26        & 8807           \\
                                                                       & FLAP       & 5.4B    & 32      & 26.9{\footnotesize±7.5}  & 8577.4{\footnotesize±2078.4}  \\
                                                                       & LLM-Pruner & 5.4B    & 32      & 24        & 8256           \\ \cline{2-6} 
                                                                       & Ours       & 5.5B    & 26      & 32        & 11008          \\ \hline
\multirow{4}{*}{\begin{tabular}[c]{@{}c@{}}27\%\\ Pruned\textsuperscript{$\dagger$}\end{tabular}} & Wanda-sp   & 4.9B    & 32      & 23        & 7816           \\
                                                                       & FLAP       & 4.9B    & 32      & 24.6{\footnotesize±8.6}  & 7497.1{\footnotesize±2358.0}  \\
                                                                       & LLM-Pruner & 4.9B    & 32      & 21        & 7155           \\ \cline{2-6} 
                                                                       & Ours       & 4.9B    & 23      & 32        & 11008          \\ \hline
\multirow{4}{*}{\begin{tabular}[c]{@{}c@{}}35\%\\ Pruned\textsuperscript{$\dagger$}\end{tabular}} & Wanda-sp   & 4.5B    & 32      & 21        & 7156           \\
                                                                       & FLAP       & 4.5B    & 32      & 23.0{\footnotesize±8.8}  & 6781.1{\footnotesize±2440.6}  \\
                                                                       & LLM-Pruner & 4.4B    & 32      & 18        & 6054           \\ \cline{2-6} 
                                                                       & Ours       & 4.5B    & 21      & 32        & 11008          \\ \hline
\multirow{4}{*}{\begin{tabular}[c]{@{}c@{}}45\%\\ Pruned\textsuperscript{$\dagger$}\end{tabular}} & Wanda-sp   & 3.7B    & 32      & 17        & 5835           \\
                                                                       & FLAP       & 3.7B    & 32      & 18.9{\footnotesize±8.0}  & 5506.8{\footnotesize±2444.7}  \\
                                                                       & LLM-Pruner & 3.7B    & 32      & 14        & 4513           \\ \cline{2-6} 
                                                                       & Ours       & 3.7B    & 17      & 32        & 11008          \\ \hline
\multirow{4}{*}{\begin{tabular}[c]{@{}c@{}}60\%\\ Pruned\textsuperscript{$\dagger$}\end{tabular}} & Wanda-sp   & 2.7B    & 32      & 12        & 4128           \\
                                                                       & FLAP       & 2.7B    & 32      & 12.7{\footnotesize±5.2}  & 4083.6{\footnotesize±2359.1}  \\
                                                                       & LLM-Pruner & 2.7B    & 32      & 8         & 2421           \\ \cline{2-6} 
                                                                       & Ours       & 2.7B    & 12      & 32        & 11008          \\ \hline
\multirow{4}{*}{\begin{tabular}[c]{@{}c@{}}80\%\\ Pruned\textsuperscript{$\dagger$}\end{tabular}} & Wanda-sp   & 1.5B    & 32      & 6         & 2059           \\
                                                                       & FLAP       & 1.5B    & 32      & 6.7{\footnotesize±2.5}   & 1988.2{\footnotesize±852.0}   \\
                                                                       & LLM-Pruner & 1.5B    & 32      & 1         & 11             \\ \cline{2-6} 
                                                                       & Ours       & 1.5B    & 6       & 32        & 11008          \\ 
\specialrule{.2em}{.1em}{.1em} 
\specialrule{.2em}{.1em}{.1em} 
                                                                       
\multicolumn{2}{c|}{Model}                                                          & \#Param & \#Block\textsuperscript{$\ddagger$} & \#Head\textsuperscript{$\ddagger$}    & FFN-D\textsuperscript{$\ddagger$}          \\ \hline
\multicolumn{2}{c|}{Original 13B}                                                   & 13.0B   & 40      & 40        & 13824          \\ \hline
\multirow{4}{*}{\begin{tabular}[c]{@{}c@{}}21\%\\ Pruned\textsuperscript{$\dagger$}\end{tabular}}                                                 & Wanda-sp   & 10.5B   & 40      & 32        & 11060          \\
                                                                       & FLAP       & 10.5B   & 40      & 33.7{\footnotesize±8.9}  & 10778.7{\footnotesize±2316.0} \\
                                                                       & LLM-Pruner & 10.3B   & 40      & 30        & 10368          \\ \cline{2-6} 
                                                                       & Ours       & 10.5B   & 32      & 40        & 13824          \\ \hline
\multirow{4}{*}{\begin{tabular}[c]{@{}c@{}}29\%\\ Pruned\textsuperscript{$\dagger$}\end{tabular}}                                                    & Wanda-sp   & 9.5B    & 40      & 29        & 9954           \\
                                                                       & FLAP       & 9.5B    & 40      & 31.1{\footnotesize±10.6} & 9570.8{\footnotesize±2601.0}  \\
                                                                       & LLM-Pruner & 9.2B    & 40      & 26        & 8985           \\ \cline{2-6} 
                                                                       & Ours       & 9.5B    & 29      & 40        & 13824          \\ \hline
\multirow{4}{*}{\begin{tabular}[c]{@{}c@{}}37\%\\ Pruned\textsuperscript{$\dagger$}\end{tabular}}                                                  & Wanda-sp   & 8.4B    & 40      & 26        & 8710           \\
                                                                       & FLAP       & 8.3B    & 40      & 27.5{\footnotesize±11.3} & 8326.6{\footnotesize±2874.9}  \\
                                                                       & LLM-Pruner & 8.2B    & 40      & 22        & 7603           \\ \cline{2-6} 
                                                                       & Ours       & 8.3B    & 25      & 40        & 13824          \\
\specialrule{.2em}{.1em}{.1em} 
\end{tabular}

\begin{tablenotes}[para,flushleft]
\footnotesize 
\textsuperscript{$\dagger$}Reduction ratio for the number of parameters.
\newline
\textsuperscript{$\ddagger$}\#Block: \#Transformer blocks; \#Head: \#attention heads of MHA; FFN-D: intermediate size of FFN. 
\end{tablenotes}
\end{threeparttable}
\end{adjustbox}

\caption{Pruned architectures on LLaMA-7B and Vicuna-\{7B, 13B\}-v1.3. While Wanda-sp~\cite{wanda,flap}, FLAP~\cite{flap}, and LLM-Pruner~\cite{llmpruner} reduce the network width, our method reduces the network depth. For moderate pruning ratios under 40\%, we used the parameter numbers from LLM-Pruner's module-level removal ratios of 25\%, 35\%, and 45\% as references and adjusted the pruning ratios for our method and the other baselines.}

\label{supple_table:arch}
\end{table}

\end{document}